\documentclass{article}

\usepackage{arxiv}

\usepackage[utf8]{inputenc} % allow utf-8 input
\usepackage[T1]{fontenc}    % use 8-bit T1 fonts
\usepackage{url}            % simple URL typesetting
\usepackage{booktabs}       % professional-quality tables
\usepackage{amsfonts}       % blackboard math symbols
\usepackage{amsmath}
\usepackage{nicefrac}       % compact symbols for 1/2, etc.
\usepackage{microtype}      % microtypography
\usepackage{graphicx}
\usepackage[table]{xcolor}
\usepackage{tabularx}
\usepackage{makecell}
\usepackage{longtable}      % Table S8
\usepackage[section]{placeins} % keep floats within their section
\usepackage{fancyvrb}       % request and prompt templates in Section S2
\usepackage[font=small,labelfont=bf,labelsep=period]{caption}
\usepackage[numbers,square,sort&compress]{natbib}
\usepackage{hyperref}       % hyperlinks
\usepackage{doi}

\hypersetup{colorlinks=true, linkcolor=blue!50!black, citecolor=blue!50!black, urlcolor=blue!50!black}

\newcolumntype{L}{>{\raggedright\arraybackslash}X}

\setcellgapes{1pt}
\newcommand{\tabnote}[1]{\par\smallskip\begin{minipage}{\linewidth}\footnotesize\raggedright #1\end{minipage}}
\newcommand{\panel}[1]{\par\smallskip{\small\textbf{#1}}\par\smallskip}
\definecolor{keygreen}{HTML}{C6EFCE}
\newcommand{\keyopt}[1]{\par\noindent{\setlength{\fboxsep}{1pt}\colorbox{keygreen}{\parbox{\dimexpr\linewidth-2\fboxsep\relax}{\raggedright #1}}}\par}

\title{Decision and System One models in rheumatology: accuracy and calibration on specialty-training examination questions and diagnostic case vignettes}

\newcommand{\orcid}[1]{\href{https://orcid.org/#1}{\includegraphics[scale=0.06]{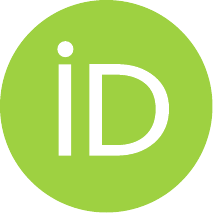}}\hspace{1mm}}

\author{%
  \orcid{0000-0002-1591-0467}Alfredo Madrid-Garc\'{\i}a\textsuperscript{1,*} \And
  \orcid{0000-0001-9119-5330}Diego Benavent\textsuperscript{2} \And
  \orcid{0009-0003-9200-477X}Fabian Lechner\textsuperscript{3} \AND
  \orcid{0009-0009-9897-2302}Phillip Kremer\textsuperscript{4} \And
  \orcid{0000-0001-9695-0657}Johannes Knitza\textsuperscript{3} \And
  \orcid{0000-0001-5070-4178}Beatriz Merino-Barbancho\textsuperscript{1}
}

\date{%
  \small
  \textsuperscript{1}Independent researcher, Madrid, Spain\\
  \textsuperscript{2}Rheumatology Department, Hospital Universitari de Bellvitge, Barcelona, Spain\\
  \textsuperscript{3}Institute for Digital Medicine, Marburg University, Marburg, Germany\\
  \textsuperscript{4}Division of Rheumatology and Systemic Inflammatory Disorders,\\ University Medical Center Hamburg-Eppendorf, Hamburg, Germany\\[4pt]
  \textsuperscript{*}Corresponding author: Alfredo Madrid-Garc\'{\i}a, Madrid, Spain. E-mail: \href{mailto:fredymad@msn.com}{\texttt{fredymad@msn.com}}%
}

\renewcommand{\headeright}{A Preprint}
\renewcommand{\undertitle}{A Preprint}
\renewcommand{\shorttitle}{System One models in rheumatology}

\hypersetup{
  pdftitle={Decision and System One models in rheumatology: accuracy and calibration on specialty-training examination questions and diagnostic case vignettes},
  pdfsubject={cs.CL, cs.AI},
  pdfauthor={Alfredo Madrid-García, Diego Benavent, Fabian Lechner, Phillip Kremer, Johannes Knitza, Beatriz Merino-Barbancho},
  pdfkeywords={Rheumatology, Artificial intelligence, Large language models, System One models, Calibration, Differential diagnosis, Medical education, Jev},
}

\begin{document}
\maketitle

\begin{abstract}
\textbf{Objective:} To evaluate the accuracy and calibration of Decision and System One models on rheumatology examination questions and diagnostic vignettes.

\textbf{Methods:} Two benchmarks were analysed: 172 text-based rheumatology questions from Spanish specialty-training examinations (expanded RheumaMIR dataset) in Spanish and English, and 70 English diagnostic vignettes from RheumBench, each with 5--18 candidate diagnoses, 52 of which included the correct reference diagnosis. Four System One models (i.e., Jev 1.13, Clef, Microsoft-Decision-1, Laya) and GPT-6 Luna, evaluated as a decision model through the OpenAI Decisions API, were compared with GPT-6 Sol, a frontier generative model, and two local generative models (Gemma 4 12B, Qwen3-8B). The primary outcome was accuracy, the proportion of questions or cases in which the selected option matched the answer key; secondary outcomes included calibration, discrimination, accuracy of answers with probability $\geq$0.90 and, in RheumaMIR, language differences.

\textbf{Results:} In RheumaMIR, Jev's accuracy was 93.0\% in Spanish and 91.9\% in English, below GPT-6 Sol (99.4\% in both languages; Holm-adjusted P$\leq$0.005); GPT-6 Luna reached 97.7\% in both languages, Clef 91.9\% and 93.0\%, and Microsoft-Decision-1 87.8\% and 86.0\%, respectively, without detectable language differences. Jev's answers with probability $\geq$0.90 (76\% and 77\% of answers) were 98.5\% correct. In RheumBench, accuracy was lower for every model: GPT-6 Sol 76.9\%, GPT-6 Luna 63.5\%, Clef 59.6\%, Jev 50.0\% and Microsoft-Decision-1 40.4\% (each System One model below GPT-6 Sol, Holm-adjusted P$\leq$0.045), against 9.6\% expected from random guessing. Jev, whose probabilities matched its accuracy in RheumaMIR (over/underconfidence index $-$0.015 and 0.000), became overconfident (mean probability 0.659; index +0.159), and its answers with probability $\geq$0.90 (25\% of cases) were 84.6\% correct, against 95.5\% for GPT-6 Luna (42\% of cases). Laya performed near chance in both benchmarks.

\textbf{Conclusion:} The best-performing System One models achieved high accuracy on rheumatology examination questions, but their performance declined on diagnostic vignettes and remained below GPT-6 Sol. Jev's high-confidence predictions were less reliable on diagnostic vignettes than on examination questions, highlighting the need for task-specific evaluation and improved calibration of diagnostic probabilities.
\end{abstract}


\paragraph{Key messages}\mbox{}
\begin{itemize}
\item Jev retained three quarters of examination answers with 98.5\% accuracy at a 0.90 probability threshold.

\item All models were less accurate than GPT-6 Sol on diagnostic vignettes.

\item No model showed a practical calibration advantage over GPT-6 Sol. Their probabilities need validation for each task before clinical use.
\end{itemize}

\section{Introduction}\label{sec:intro}

Large language models (LLMs) have been evaluated extensively in rheumatology, on tasks ranging from examination questions and differential diagnosis to information extraction and education~\citep{Benavent2025,Krusche2024}. In the RheumaMIR study~\citep{MadridGarcia2023}, GPT-4 correctly answered 93.7\% of the rheumatology questions from the Spanish access examination to specialised medical training. In RheumBench, a specialist-validated benchmark of diagnostic and management cases, the best of ten systems ranked the correct diagnosis first in 57.1\% of 70 diagnostic cases answered in free text~\citep{Lechner2026}. However, because LLMs generate their answers token by token, they have limitations that matter in clinical practice. They can produce plausible but incorrect content~\citep{Asgari2025}. Their text outputs may need parsing or normalisation before automated use, which can introduce formatting and extraction errors~\citep{Grothey2025,Naderi2026}. The numerical confidence they report tends to be overestimated and discriminates poorly between correct and incorrect answers~\citep{Savage2025,Naderi2026}; for instance, in RheumBench, the confidence stated by every system exceeded its diagnostic accuracy~\citep{Lechner2026}. Longer sequential outputs also increase generation latency and inference cost, which is relevant for deployment at health-system scale~\citep{Bhattacharyya2026}.

Nevertheless, many clinical applications of artificial intelligence (AI) need a choice among predefined alternatives rather than free text, such as deciding whether a patient meets trial eligibility criteria or whether an article meets review inclusion criteria~\citep{Jin2024,Unlu2025,Guo2024,Sanghera2025}. Much of rheumatology decision-making likewise consists of structured judgements based on unstructured information. Classification criteria, for example, are weighted sums of explicitly defined items~\citep{Aletaha2010}, and routine work includes triaging referrals, checking requirements before biologic therapy and coding diagnoses. Each of these judgements could be posed as a question with predefined answers. The probability assigned to each answer could then be used to select which answers a rheumatologist reviews. Such tasks require an answer restricted to the predefined options, with a probability that reflects how likely it is to be correct.

\begin{figure}[p]
\centering
\includegraphics[width=\linewidth,height=0.86\textheight,keepaspectratio]{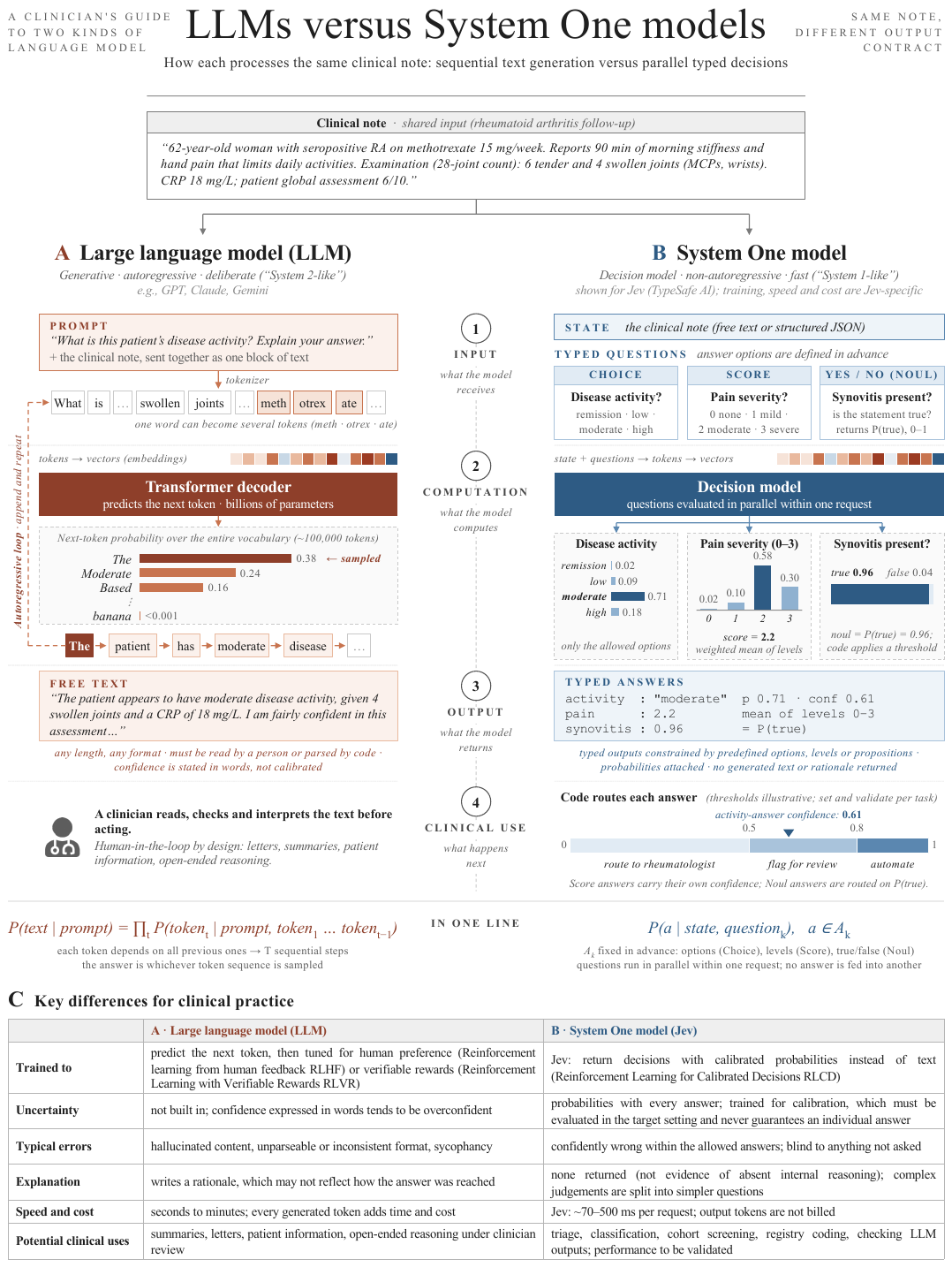}
\caption{How a large language model (LLM) and a System One model process the same clinical note. Diagram drafted with large language model assistance; content verified by the authors.}\label{fig:1}
\end{figure}

Decision and System One models are a recently introduced class of AI models designed for this kind of decision (Figure~\ref{fig:1})~\citep{Almeida2026,TypeSafeAI2026docs}. They receive a ``state'' describing a case, along with typed questions whose possible answers are defined in advance, and return a probability for each allowed answer. The output is machine-readable and cannot fall outside the options provided, but it gives no rationale that a clinician could inspect.

Within one month of Jev's release, the first popularised model of this class, more than 100 decision models had been listed on independent leaderboards, including solutions from OpenAI, Microsoft, AWS and Cloudflare~\citep{BenchmarkHeaven2026}. Open-source alternatives include Laya~\citep{ConvaiInnovations2026} and Clef~\citep{Chen2026}.

Evidence on System One models in medicine is scarce. In a benchmark of Jev on four medical datasets, its accuracy was similar to that of a frontier reasoning LLM, GPT-6 Sol, on questions about research abstracts but lower on examination questions and complex diagnostic cases, and its better-calibrated probabilities did not yield more answers that could be accepted at a given accuracy~\citep{MadridGarcia2026jev}.

To our knowledge, no System One model has been evaluated in rheumatology or in a language other than English. Notably, Jev's primary training language is English~\citep{TypeSafeAI2026docs}. Likewise, the application of these models to complex diagnostic cases, where the correct diagnosis must be chosen among many plausible alternatives and may be absent from the options offered has not been studied.

We therefore evaluated four System One models (Jev, Clef, Microsoft-Decision-1 and Laya) and GPT-6 Luna through OpenAI's Decisions API on RheumaMIR and RheumBench. We assessed how often each model selected the official answer, whether its probabilities identified answers more likely to be correct, and whether accuracy differed between Spanish and English.

\section{Methods}\label{sec:methods}

\subsection{Study design}\label{sec:design}

This was a cross-sectional benchmark study of nine model conditions on RheumaMIR and RheumBench. No participants were recruited and no identifiable patient data were used. The study is reported following the TRIPOD-LLM guideline~\citep{Gallifant2025}. In line with recommendations for LLM-based rheumatology research~\citep{MadridGarcia2026scoping}, we report model versions, openness, access route and evaluation dates, together with the code, requests and responses (see Data availability). For brevity, GPT-6 Luna is grouped with the System One models in the tables and comparisons.

\subsection{Datasets}\label{sec:datasets}

\begin{itemize}
\item \textbf{RheumaMIR}~\citep{MadridGarcia2023}. The MIR is the annual national examination for access to specialised medical training in Spain, with single-best-answer questions (four to five options) and definitive answer keys published by the Ministry of Health. RheumaMIR contains 145 rheumatology-related records from the 2009--2010 to 2022--2023 examinations; image-dependent questions were excluded when the dataset was built. We added 30 text-based rheumatology-related questions from the three subsequent examinations (2023--2024 to 2025--2026), selected by a rheumatologist, with keys checked against the definitive answer tables of the Ministry. After exclusion of two duplicate records and of one question invalidated by the Ministry, 172 questions were analysed. Each question was evaluated in Spanish and in English. The English versions had been translated automatically. All English translations were then reviewed against the Spanish with Claude Opus 5.5 (Anthropic), which was also used to correct language errors in the Spanish text. Further details can be found in Supplementary Section~\ref{sec:S1.1} and Table~\ref{tab:S1}.

\item \textbf{RheumBench}~\citep{Lechner2026}. RheumBench is a specialist-validated benchmark for evaluating LLMs in rheumatology. Its 70 diagnostic cases comprise 60 published case reports of rare rheumatic diseases and clinically important differential diagnoses, and 10 de-identified cases from the University Hospital Hamburg; 42 represent inflammatory rheumatic diseases and 28 non-inflammatory differential diagnoses. In the original benchmark, systems proposed diagnoses in free text that were scored against case-specific rubrics. For this study, each vignette was paired with 5 to 18 candidate diagnoses in alphabetical order proposed by systems investigated in the RheumBench study. In 52 cases one candidate names the reference diagnosis; in the other 18, none does. Only the option naming the reference diagnosis was scored as correct. In 12 cases the option that words the reference diagnosis differently was matched by hand. More details can be found in Supplementary Section~\ref{sec:S1.2}.
\end{itemize}

\subsection{Models}\label{sec:models}

The evaluated models, how they were accessed and the considerations relevant to the interpretation of their results are summarised in Supplementary Table~\ref{tab:S2}. Four System One models (Jev 1.13, Clef, Microsoft-Decision-1 and Laya) and GPT-6 Luna, accessed through OpenAI's Decisions API, were evaluated. GPT-6 Sol with medium reasoning was the primary comparator; GPT-6 Sol without reasoning and two generative models run locally (Gemma 4 12B and Qwen3-8B) were secondary comparators. No model was fine-tuned.

System One models and GPT-6 Luna return a probability for each option as a direct output, without generating text. The generative models were asked in the prompt to write a probability for each option; these stated probabilities are self-reported estimates.

\subsection{Evaluation procedure}\label{sec:procedure}

In RheumaMIR, all models were provided with the question stem, the response options in their original numbering and order, and a neutral instruction in the language of the item (e.g., ``\emph{Which option is the correct answer to this question?}''). System One models received the stem and the options as a single-choice question, and generative models received them in one user message with a request for the answer number and a probability for each option (Supplementary Section~\ref{sec:S2}). No system prompt, example, or reference answer was used, and prompts were not optimised. Requests were sent between 5 and 10 October 2026. The answer selected by the model was scored as returned, even when another option had a higher probability. The local generative models, whose server could not enforce an output format, were asked to answer only with a JSON object, and the first JSON object in each response was scored.

In RheumBench, every model received the instruction ``\emph{Which option is the most likely diagnosis?}'', the vignette verbatim and the numbered candidate diagnoses, in the same request formats as in RheumaMIR; generative models received the vignette under the label ``\emph{Case:}''. In the primary option set, the candidate diagnoses were given as listed, and all 70 cases were sent. In a secondary option set, ``\emph{None of the above}'' was added as the last option of every case, so that it was the correct option in the 18 cases without the reference diagnosis and all 70 cases had a correct option.

Every condition was run three times with identical requests in each dataset and option set. The first run was the primary analysis; later runs assessed repeatability. Repeated runs and language versions were paired observations of the same items, not independent samples.

\subsection{Outcomes and statistical analysis}\label{sec:stats}

The question or case was the unit of analysis. The primary outcome was accuracy, the proportion of items for which the selected option matched the reference answer, with Wilson 95\% confidence intervals (CIs). Paired comparisons used the exact McNemar test, with CIs for differences from 2,000 bootstrap resamples of items that kept paired outputs together. Holm adjustment covered three families comprising ten System One versus GPT-6 Sol medium comparisons across both RheumaMIR languages, five System One language comparisons, and five System One versus GPT-6 Sol medium comparisons in RheumBench (Supplementary Section~\ref{sec:S3} and Table~\ref{tab:S3}). Secondary comparators were described without significance tests. Accuracy expected from random guessing was the mean reciprocal of the number of options. Agreement between languages was summarised as the percentage of questions with the same selected option and Cohen's kappa. For Laya, which used different checkpoints for English and Spanish items (Supplementary Table~\ref{tab:S2}), comparisons between languages also reflect the checkpoint.

Probability analyses used the probability of the selected option (selected-answer probability); System One probabilities, including those returned by the Decisions API, were analysed as returned and stated probabilities were rescaled to sum to 1 over the offered options, keeping equal stated probabilities tied (Supplementary Section~\ref{sec:S2}). Calibration was summarised by the over/underconfidence index, with reliability diagrams (Figure~\ref{fig:2} and Supplementary Figure~\ref{fig:S1}) and with the expected calibration error (ECE; ten equal-width bins closed on the right; Supplementary Section~\ref{sec:S4}), reported descriptively without CIs. Discrimination between correct and incorrect answers was summarised by the correctness AUROC, with bootstrap CIs (Supplementary Table~\ref{tab:S4}). To assess a possible review strategy, answers with a selected-answer probability of at least 0.90 were retained and the rest deferred; we report the number and proportion of answers retained (coverage) and the accuracy of the retained answers.

In both benchmarks, repeatability was summarised as the percentage of items with the same answer in all three runs and the range of single-run accuracy (Supplementary Table~\ref{tab:S5}). Sensitivity analyses recomputed the ECE with five equal-width bins, scored the highest-probability option instead of the returned answer and, for Laya, restricted RheumBench to the vignettes it read in full (Supplementary Section~\ref{sec:S4}). Latency and cost were described in Supplementary Table~\ref{tab:S6}.

In RheumBench, accuracy, probability measures and retained answers were analysed in the 52 cases whose reference diagnosis was among the options. The other 18 were analysed as an exploratory forced-choice test, counting the forced answers that the 0.90 threshold would have accepted. In the None of the above option set, we report how often None of the above was selected in the 18 and in the 52 cases, and accuracy over the 70 cases (Supplementary Table~\ref{tab:S7}).

To describe the most confident errors, we listed the ten RheumaMIR questions answered incorrectly with the highest selected-answer probabilities by Jev, GPT-6 Luna, Clef, Microsoft-Decision-1 or GPT-6 Sol with medium reasoning (Supplementary Table~\ref{tab:S8}).

Analyses used Python 3.13 with a fixed random seed.

\section{Results}\label{sec:results}

\subsection{RheumaMIR: accuracy and language}\label{sec:res-mir-acc}

Of 172 questions per language, Jev answered 160 (93.0\%) correctly in Spanish and 158 (91.9\%) in English, versus 171 (99.4\%) for GPT-6 Sol in both languages, with and without reasoning (Table~\ref{tab:1}). Paired values below follow this language order. All differences refer to GPT-6 Sol with medium reasoning, and all P values are Holm-adjusted (Supplementary Table~\ref{tab:S3}). Jev was 6.4 and 7.6 percentage points less accurate (P=0.005 and P=0.001), and GPT-6 Sol answered all but one of Jev's errors correctly in each language. GPT-6 Sol's only discordance with the answer key, a 2011--2012 hip-fracture prevention question, was shared by Jev, GPT-6 Luna, Microsoft-Decision-1, Qwen3-8B and, in English, Laya (Supplementary Sections~\ref{sec:S1.1} and~\ref{sec:S3}). GPT-6 Luna answered 168 (97.7\%) correctly in both languages ($-$1.7 percentage points; P=0.500).

\begin{table}[tbp]
\caption{Accuracy by model (primary outcome).}\label{tab:1}
\centering\small
\setlength{\tabcolsep}{4pt}
\panel{(A) RheumaMIR (172 questions per language)}
\begin{tabular}{@{}lcc>{\centering\arraybackslash}p{2.7cm}c@{}}
\toprule
Model & \makecell[b]{Spanish, n/N;\\\% (95\% CI)} & \makecell[b]{English, n/N;\\\% (95\% CI)} & \makecell[b]{English $-$ Spanish,\\pp (95\% CI)} & \makecell[b]{P\\(Holm)\textsuperscript{a}}\\
\midrule
\multicolumn{5}{@{}l}{\textbf{\emph{System One models}}}\\
\quad Jev 1.13 & 160/172; 93.0 (88.2--96.0) & 158/172; 91.9 (86.8--95.1) & $-$1.2 ($-$5.2 to 2.3) & 1.000\\
\quad GPT-6 Luna (Decisions API) & 168/172; 97.7 (94.2--99.1) & 168/172; 97.7 (94.2--99.1) & 0.0 ($-$2.3 to 2.3) & 1.000\\
\quad Clef & 158/172; 91.9 (86.8--95.1) & 160/172; 93.0 (88.2--96.0) & +1.2 ($-$2.9 to 5.2) & 1.000\\
\quad Microsoft-Decision-1 & 151/172; 87.8 (82.1--91.9) & 148/172; 86.0 (80.1--90.4) & $-$1.7 ($-$6.4 to 2.3) & 1.000\\
\quad Laya 0.3.20 & 38/172; 22.1 (16.5--28.9) & 43/172; 25.0 (19.1--32.0) & +2.9 ($-$5.2 to 11.0) & 1.000\\
\addlinespace[3pt]
\multicolumn{5}{@{}l}{\textbf{\emph{Primary comparator}}}\\
\quad GPT-6 Sol (medium) & 171/172; 99.4 (96.8--99.9) & 171/172; 99.4 (96.8--99.9) & 0.0 (no discordant answers) & --\\
\addlinespace[3pt]
\multicolumn{5}{@{}l}{\textbf{\emph{Secondary comparators}}}\\
\quad GPT-6 Sol (none) & 171/172; 99.4 (96.8--99.9) & 171/172; 99.4 (96.8--99.9) & 0.0 (no discordant answers) & --\\
\quad Gemma 4 12B (4-bit, no thinking) & 139/172; 80.8 (74.3--86.0) & 142/172; 82.6 (76.2--87.5) & +1.7 ($-$2.9 to 7.0) & --\\
\quad Qwen3-8B (4-bit, no thinking) & 110/172; 64.0 (56.5--70.8) & 117/172; 68.0 (60.7--74.5) & +4.1 ($-$2.3 to 10.5) & --\\
\addlinespace[3pt]
Random guessing (expected) & 23.2 & 23.2 & -- & --\\
\bottomrule
\end{tabular}
\panel{(B) RheumBench (52 cases whose reference diagnosis is among the options)}
\begin{tabular}{@{}lccc@{}}
\toprule
Model & \makecell[b]{Accuracy, n/52;\\\% (95\% CI)} & \makecell[b]{Model $-$ GPT-6 Sol (medium),\\pp (95\% CI)} & \makecell[b]{P\\(Holm)\textsuperscript{a}}\\
\midrule
\multicolumn{4}{@{}l}{\textbf{\emph{System One models}}}\\
\quad Jev 1.13 & 26/52; 50.0 (36.9--63.1) & $-$26.9 ($-$42.3 to $-$9.6) & 0.013\\
\quad GPT-6 Luna (Decisions API) & 33/52; 63.5 (49.9--75.2) & $-$13.5 ($-$25.0 to $-$3.8) & 0.045\\
\quad Clef & 31/52; 59.6 (46.1--71.8) & $-$17.3 ($-$30.8 to $-$3.8) & 0.045\\
\quad Microsoft-Decision-1 & 21/52; 40.4 (28.2--53.9) & $-$36.5 ($-$51.9 to $-$23.1) & \textless{}0.001\\
\quad Laya 0.3.20 & 5/52; 9.6 (4.2--20.6) & $-$67.3 ($-$78.8 to $-$53.8) & \textless{}0.001\\
\addlinespace[3pt]
\multicolumn{4}{@{}l}{\textbf{\emph{Primary comparator}}}\\
\quad GPT-6 Sol (medium) & 40/52; 76.9 (63.9--86.3) & -- & --\\
\addlinespace[3pt]
\multicolumn{4}{@{}l}{\textbf{\emph{Secondary comparators}}}\\
\quad GPT-6 Sol (none) & 39/52; 75.0 (61.8--84.8) & -- & --\\
\quad Gemma 4 12B (4-bit, no thinking) & 21/52; 40.4 (28.2--53.9) & -- & --\\
\quad Qwen3-8B (4-bit, no thinking) & 16/52; 30.8 (19.9--44.3) & -- & --\\
\addlinespace[3pt]
Random guessing (expected) & 9.6 & -- & --\\
\bottomrule
\end{tabular}
\tabnote{%
Run 1; RheumBench, listed option set. Accuracy with Wilson 95\% CIs. Differences are paired, in percentage points (pp), with 95\% CIs from 2,000 bootstrap resamples of questions or cases; P values from exact McNemar tests. In RheumaMIR, every System One model except GPT-6 Luna was less accurate than GPT-6 Sol with medium reasoning in both languages (Holm-adjusted P$\leq$0.010, ten comparisons); GPT-6 Luna was 1.7 percentage points less accurate in each language (Holm-adjusted P=0.500) (Supplementary Table~\ref{tab:S3}). Secondary comparators were not tested (--). GPT-6 Sol (medium) and (none): medium and no reasoning effort. Gemma 4 12B and Qwen3-8B (4-bit, no thinking): 4-bit conversions run locally with thinking disabled. Laya answered English and Spanish items with different checkpoints, so its language comparison is also a checkpoint comparison.
\par\smallskip
\textsuperscript{a} Holm-adjusted within each family of five System One comparisons: English versus Spanish (A) and each model versus GPT-6 Sol with medium reasoning (B).
}
\end{table}

Clef answered 158 (91.9\%) and 160 (93.0\%) correctly ($-$7.6 and $-$6.4 percentage points; P=0.005 and P=0.010) and Microsoft-Decision-1 151 (87.8\%) and 148 (86.0\%) ($-$11.6 and $-$13.4 percentage points; both P\textless{}0.001). Laya performed near chance (22.1\% and 25.0\%, against 23.2\% expected). Gemma 4 12B (80.8\% and 82.6\%) and Qwen3-8B (64.0\% and 68.0\%) were less accurate than every remote System One model. No System One model showed a statistically detectable difference in accuracy between languages (all Holm-adjusted P=1.000), although the CIs did not exclude differences of several percentage points (Jev, $-$1.2; 95\% CI $-$5.2 to 2.3). Jev selected the same option in both languages for 94.2\% of questions (Cohen's kappa 0.92), GPT-6 Luna for 97.7\% (0.97), Clef for 91.9\% (0.89) and Microsoft-Decision-1 for 89.5\% (0.86), against 34.9\% for Laya. Accuracy in the 30 new questions was within 4 percentage points of that in the historical questions for Jev, GPT-6 Luna and Clef, and 3--9 points lower for Microsoft-Decision-1.

\subsection{RheumaMIR: probabilities and retained answers}\label{sec:res-mir-prob}

\begin{figure}[tbp]
\centering
\panel{(A)}
\includegraphics[width=\linewidth]{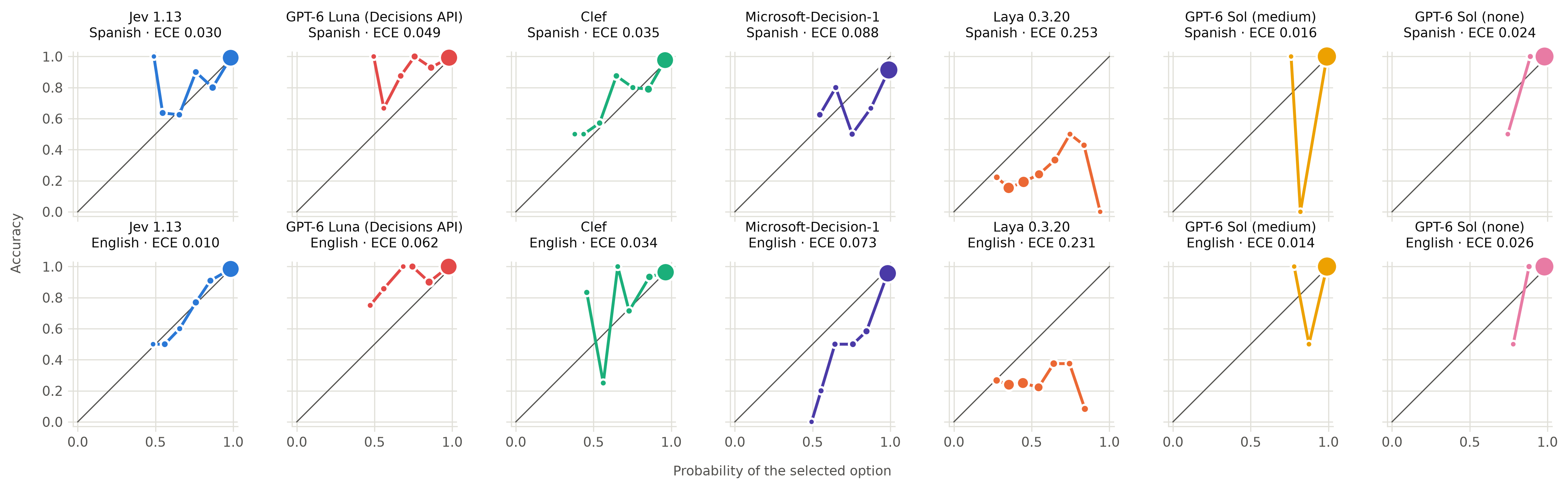}
\panel{(B)}
\includegraphics[width=\linewidth]{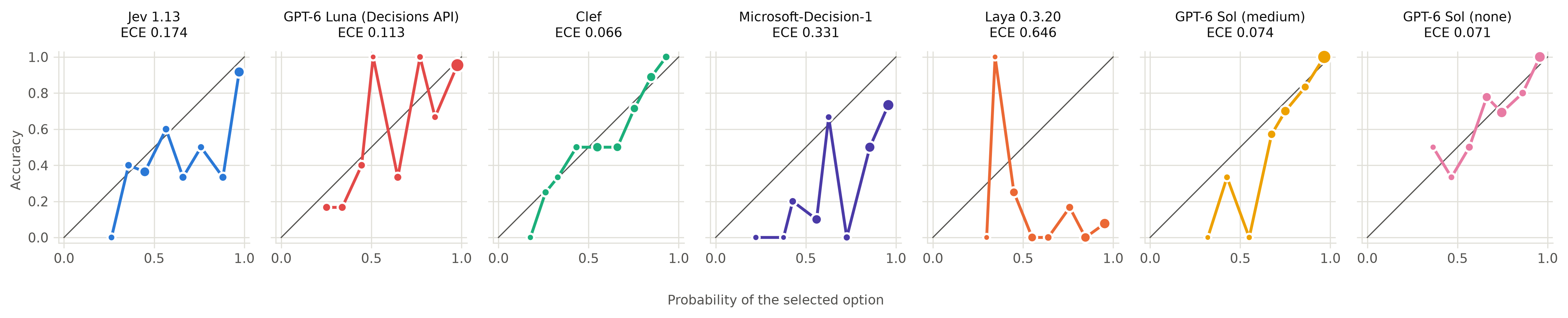}
\caption{Reliability diagrams of the System One models and GPT-6 Sol (run 1). (A) RheumaMIR, Spanish (top row) and English (bottom row), 172 questions per language. (B) RheumBench, listed option set, 52 cases whose reference diagnosis is among the options. Each point is one of ten equal-width bins (closed on the right) of the selected-answer probability, plotted against the accuracy of those answers; point area is proportional to the number of answers; points below the diagonal indicate overconfidence; panel titles give the ECE. GPT-6 Sol is shown with medium and with no reasoning.}\label{fig:2}
\end{figure}

Jev's probabilities matched its accuracy (over/underconfidence index $-$0.015 and 0.000), had low calibration error (ECE 0.030 in Spanish and 0.010 in English; Table~\ref{tab:2}; Figure~\ref{fig:2}A) and separated correct from incorrect answers well (correctness AUROC 0.904 and 0.916; Supplementary Table~\ref{tab:S4}). At the 0.90 threshold, Jev retained 130 (75.6\%) and 132 (76.7\%) of the 172 answers, with 98.5\% accuracy in both languages (95\% CI 94.6--99.6). GPT-6 Luna's mean selected-answer probability was below observed accuracy (0.928 and 0.914 versus 97.7\%; index $-$0.049 and $-$0.062), and the threshold retained 135 (78.5\%) and 127 (73.8\%) answers with 99.3\% and 100.0\% accuracy.

\begin{table}[tbp]
\caption{Probability quality and answers retained at a selected-answer probability of at least 0.90.}\label{tab:2}
\centering\small
\setlength{\tabcolsep}{4pt}
\panel{(A) RheumaMIR (172 questions per language)}
\begin{tabular}{@{}lcccccc@{}}
\toprule
Model, language & \makecell[b]{Errors,\\n} & \makecell[b]{Mean\\probability\textsuperscript{a}} & \makecell[b]{O/U\\index} & ECE & \makecell[b]{Retained,\\n (\%)} & \makecell[b]{Accuracy of retained,\\\% (95\% CI)}\\
\midrule
\multicolumn{7}{@{}l}{\textbf{\emph{System One models}}}\\
\quad Jev 1.13, Spanish & 12 & 0.915 & $-$0.015 & 0.030 & 130 (75.6) & 98.5 (94.6--99.6)\\
\quad Jev 1.13, English & 14 & 0.919 & 0.000 & 0.010 & 132 (76.7) & 98.5 (94.6--99.6)\\
\quad GPT-6 Luna (Decisions API), Spanish & 4 & 0.928 & $-$0.049 & 0.049 & 135 (78.5) & 99.3 (95.9--99.9)\\
\quad GPT-6 Luna (Decisions API), English & 4 & 0.914 & $-$0.062 & 0.062 & 127 (73.8) & 100.0 (97.1--100.0)\\
\quad Clef, Spanish & 14 & 0.898 & $-$0.021 & 0.035 & 129 (75.0) & 97.7 (93.4--99.2)\\
\quad Clef, English & 12 & 0.913 & $-$0.017 & 0.034 & 137 (79.7) & 96.4 (91.7--98.4)\\
\quad Microsoft-Decision-1, Spanish & 21 & 0.950 & +0.072 & 0.088 & 150 (87.2) & 91.3 (85.7--94.9)\\
\quad Microsoft-Decision-1, English & 24 & 0.934 & +0.073 & 0.073 & 138 (80.2) & 95.7 (90.8--98.0)\\
\quad Laya 0.3.20, Spanish & 134 & 0.474 & +0.253 & 0.253 & 1 (0.6) & 0.0 (0.0--79.3)\\
\quad Laya 0.3.20, English & 129 & 0.481 & +0.231 & 0.231 & 0 (0.0) & NE\\
\addlinespace[3pt]
\multicolumn{7}{@{}l}{\textbf{\emph{Primary comparator}}}\\
\quad GPT-6 Sol (medium), Spanish & 1 & 0.988 & $-$0.006 & 0.016 & 170 (98.8) & 100.0 (97.8--100.0)\\
\quad GPT-6 Sol (medium), English & 1 & 0.988 & $-$0.006 & 0.014 & 170 (98.8) & 100.0 (97.8--100.0)\\
\addlinespace[3pt]
\multicolumn{7}{@{}l}{\textbf{\emph{Secondary comparators}}}\\
\quad GPT-6 Sol (none), Spanish & 1 & 0.976 & $-$0.018 & 0.024 & 168 (97.7) & 100.0 (97.8--100.0)\\
\quad GPT-6 Sol (none), English & 1 & 0.975 & $-$0.019 & 0.026 & 167 (97.1) & 100.0 (97.8--100.0)\\
\quad Gemma 4 12B (4-bit, no thinking), Spanish & 33 & 0.854 & +0.046 & 0.072 & 81 (47.1) & 85.2 (75.9--91.3)\\
\quad Gemma 4 12B (4-bit, no thinking), English & 30 & 0.857 & +0.032 & 0.063 & 83 (48.3) & 84.3 (75.0--90.6)\\
\quad Qwen3-8B (4-bit, no thinking), Spanish & 62 & 0.619 & $-$0.020 & 0.112 & 10 (5.8) & 60.0 (31.3--83.2)\\
\quad Qwen3-8B (4-bit, no thinking), English & 55 & 0.623 & $-$0.057 & 0.124 & 7 (4.1) & 85.7 (48.7--97.4)\\
\bottomrule
\end{tabular}
\panel{(B) RheumBench (52 cases whose reference diagnosis is among the options)}
\begin{tabular}{@{}lcccccc@{}}
\toprule
Model & \makecell[b]{Errors,\\n} & \makecell[b]{Mean\\probability\textsuperscript{a}} & \makecell[b]{O/U\\index} & ECE & \makecell[b]{Retained,\\n (\%)} & \makecell[b]{Accuracy of retained,\\\% (95\% CI)}\\
\midrule
\multicolumn{7}{@{}l}{\textbf{\emph{System One models}}}\\
\quad Jev 1.13 & 26 & 0.659 & +0.159 & 0.174 & 13 (25.0) & 84.6 (57.8--95.7)\\
\quad GPT-6 Luna (Decisions API) & 19 & 0.703 & +0.068 & 0.113 & 22 (42.3) & 95.5 (78.2--99.2)\\
\quad Clef & 21 & 0.624 & +0.028 & 0.066 & 5 (9.6) & 100.0 (56.6--100.0)\\
\quad Microsoft-Decision-1 & 31 & 0.730 & +0.327 & 0.331 & 15 (28.8) & 73.3 (48.0--89.1)\\
\quad Laya 0.3.20 & 47 & 0.716 & +0.620 & 0.646 & 13 (25.0) & 7.7 (1.4--33.3)\\
\addlinespace[3pt]
\multicolumn{7}{@{}l}{\textbf{\emph{Primary comparator}}}\\
\quad GPT-6 Sol (medium) & 12 & 0.814 & +0.044 & 0.074 & 23 (44.2) & 100.0 (85.7--100.0)\\
\addlinespace[3pt]
\multicolumn{7}{@{}l}{\textbf{\emph{Secondary comparators}}}\\
\quad GPT-6 Sol (none) & 13 & 0.748 & $-$0.002 & 0.071 & 15 (28.8) & 100.0 (79.6--100.0)\\
\quad Gemma 4 12B (4-bit, no thinking) & 31 & 0.667 & +0.263 & 0.313 & 1 (1.9) & 100.0 (20.7--100.0)\\
\quad Qwen3-8B (4-bit, no thinking) & 36 & 0.609 & +0.301 & 0.301 & 0 (0.0) & NE\\
\bottomrule
\end{tabular}
\tabnote{%
Run 1; RheumBench, listed option set. Probabilities refer to the selected answer; those of the System One models were returned by their APIs or packages and used as returned, and those of the generative models were stated in response to the prompt and rescaled to sum to 1. ECE, expected calibration error with ten equal-width bins closed on the right (descriptive, without CI). O/U index, over/underconfidence index: mean probability minus accuracy, calculated from unrounded values (positive values indicate overconfidence). Retained answers had a selected-answer probability of at least 0.90; coverage is retained/172 (A) or retained/52 (B), and the accuracy of retained answers is given with Wilson 95\% CIs. The 0.90 threshold was evaluated on the same data and is not a clinically validated rule. NE, not estimable (no retained answers). Correctness AUROC is reported in Supplementary Table~\ref{tab:S4}. Laya's package warned that its probabilities for RheumBench questions with 11 or more options (29 of the 52 cases) were uncalibrated.
\par\smallskip
\textsuperscript{a} Mean probability of the selected option, to be read against accuracy (Table~\ref{tab:1}).
}
\end{table}

Clef also had low calibration error (ECE 0.035 and 0.034) with lower observed AUROC values than Jev (AUROC 0.851 and 0.789), and Microsoft-Decision-1 was overconfident (mean selected-answer probability 0.950 and 0.934, against 87.8\% and 86.0\% accuracy; index +0.072 and +0.073); their retained answers were 96.4--97.7\% and 91.3--95.7\% correct, respectively. The ten questions with the highest-probability errors (0.967--0.999) were all answered incorrectly by Microsoft-Decision-1 in Spanish, mostly on the choice of investigation or treatment; one, a negatively worded question on temporal arteritis, was also a high-confidence error of Jev and Clef in both languages (Supplementary Table~\ref{tab:S8}).

GPT-6 Sol with medium reasoning retained 170 (98.8\%) answers in each language, all correct, whereas Laya and the local generative models showed poor discrimination (AUROC 0.52--0.61; Supplementary Table~\ref{tab:S4} and Figure~\ref{fig:S1}).

\subsection{RheumBench: accuracy}\label{sec:res-rb-acc}

Of 52 cases, GPT-6 Sol answered 40 (76.9\%) correctly with medium reasoning and 39 (75.0\%) without, against 9.6\% expected from random guessing (Table~\ref{tab:1}). GPT-6 Luna answered 33 (63.5\%), Clef 31 (59.6\%), Jev 26 (50.0\%) and Microsoft-Decision-1 21 (40.4\%), 13.5 to 36.5 percentage points below GPT-6 Sol with medium reasoning (Holm-adjusted P$\leq$0.045; Supplementary Table~\ref{tab:S3}), although the differences for GPT-6 Luna and Clef rested on few discordant cases. GPT-6 Luna, Clef and Jev were not compared with each other, so their ranking was not established. Laya answered 5 (9.6\%) correctly, Gemma 4 12B 21 (40.4\%) and Qwen3-8B 16 (30.8\%).

\subsection{RheumBench: probabilities and retained answers}\label{sec:res-rb-prob}

Every model except GPT-6 Sol without reasoning was overconfident (over/underconfidence index +0.028 to +0.620; Table~\ref{tab:2}). Clef had the lowest ten-bin ECE (0.066) and, among the System One models, the index closest to zero (+0.028). GPT-6 Luna had the highest observed AUROC among the System One models (0.876), close to GPT-6 Sol with medium reasoning (0.872) (Figure~\ref{fig:2}B; Supplementary Table~\ref{tab:S4}). Jev was overconfident (mean selected-answer probability 0.659 against 50.0\% accuracy; index +0.159; ECE 0.174) and discriminated less clearly (AUROC 0.737). Microsoft-Decision-1 (mean 0.730 against 40.4\% accuracy; +0.327) and Laya (0.716 against 9.6\%; +0.620) were more overconfident, and the stated probabilities of the local generative models discriminated poorly (AUROC 0.570--0.620; Supplementary Figure~\ref{fig:S1}).

At the 0.90 threshold, GPT-6 Sol with medium reasoning retained 23 (44.2\%) of the 52 answers, all correct, and GPT-6 Luna retained 22 (42.3\%) with 95.5\% accuracy. Jev retained 13 (25.0\%) with 84.6\% accuracy, Microsoft-Decision-1 15 (28.8\%) with 73.3\%, and Clef 5 (9.6\%), all correct (Table~\ref{tab:2}).

\subsection{RheumBench: reference diagnosis not among the options}\label{sec:res-rb-nota}

In the 18 cases where no option named the reference diagnosis, every selected answer was wrong, yet the 0.90 threshold would still have accepted at least one of them for six of the nine conditions (Supplementary Table~\ref{tab:S7}). When None of the above was added, only GPT-6 Luna (4 of 18), Microsoft-Decision-1 (3) and Laya (9, but also 32 of the 52 cases with a correct option) selected it in these cases.

\subsection{Repeatability, latency, cost and sensitivity analyses}\label{sec:res-rep}

In both benchmarks, GPT-6 Luna, Clef and Laya returned the same answer in all three runs for every item, and GPT-6 Luna and Clef also returned identical probabilities. Jev changed its answer in five RheumaMIR question-language pairs and two of the 52 analysed RheumBench cases, and Microsoft-Decision-1 in six and three. GPT-6 Sol changed no RheumaMIR answer with medium reasoning and one without, but changed its answer in six RheumBench cases with medium reasoning and four without, although its accuracy with medium reasoning was the same in all three runs. The local generative models were less repeatable (Supplementary Table~\ref{tab:S5}). Median latency was 0.21--0.48 s for the remote System One models, against 1.36--6.53 s for GPT-6 Sol and 3.1--39.9 s for the local generative models, and costs per 1,000 items were US\$0.007--0.138, against US\$0.96--5.54 (Supplementary Table~\ref{tab:S6}). Scoring the highest-probability option changed accuracy by at most one item for every condition except Qwen3-8B in RheumaMIR, and Laya remained near chance on the RheumBench vignettes it read in full (Supplementary Section~\ref{sec:S4}).

\section{Discussion}\label{sec:discussion}

Across RheumaMIR and RheumBench, the System One models performed very differently. On MIR questions, Jev 1.13 answered more than nine of ten questions correctly and 98.5\% of the three quarters of its answers given a probability of at least 0.90 were correct; GPT-6 Luna came within two points of GPT-6 Sol; Clef was about as accurate as Jev, but its probabilities separated correct from incorrect answers less clearly; and Microsoft-Decision-1 was less accurate and overconfident. On RheumBench vignettes, accuracy was lower for every model, the remote System One models remained 13--37 points below GPT-6 Sol, and Jev's probabilities became overconfident. Laya performed at or near chance in both benchmarks, and no language difference was detected in RheumaMIR.

\subsection{Accuracy and the reference standard}\label{sec:disc-acc}

Jev's accuracy on RheumaMIR was higher than in the only previous evaluation of the model in medicine (74.8\% on MetaMedQA examination questions and 59.8--61.8\% on complex diagnostic cases)~\citep{MadridGarcia2026jev}, whereas on RheumBench (50.0\%) it was closer to that reported for complex diagnostic cases.

The gap with the frontier model remained in both benchmarks. In RheumaMIR, GPT-6 Sol answered almost all of Jev's errors correctly, and its only error, shared by Jev, GPT-6 Luna, Microsoft-Decision-1, Qwen3-8B and, in English, Laya, was a disagreement with a key set more than a decade ago that may no longer reflect current practice. Part of the residual error of the best models may therefore lie in the reference standard. The near-perfect performance of GPT-6 Sol and GPT-6 Luna also means that this dataset no longer discriminates between frontier models.

RheumBench did discriminate between them. Its vignettes describe rare diseases and their mimics and come with up to 18 candidate diagnoses, many of them close alternatives. These accuracies are not directly comparable with those of the original free-text RheumBench evaluation, because here the candidate list narrowed the search, the 18 cases without the reference diagnosis were analysed separately and the models were different. Over all 70 cases, GPT-6 Sol answered 57.1\% correctly with the listed options, the same proportion as the best system in the original free-text evaluation, so its higher accuracy in the 52 cases mainly reflects the exclusion of the 18 cases without the reference diagnosis. With None of the above added, it answered 61.4\% correctly (Supplementary Table~\ref{tab:S7}), but the gain came from the 52 cases with a correct listed option (43 against 40 correct), because GPT-6 Sol never selected None of the above.

GPT-6 Luna came within two points of GPT-6 Sol on examination questions and had the highest point accuracy among the System One models on RheumBench, suggesting that typed answers need not entail a large loss of accuracy; whether this reflects the model, its training or the interface cannot be determined from a black-box comparison.

Clef and Microsoft-Decision-1 show that the typed interface does not by itself determine accuracy. Clef matched Jev's accuracy in RheumaMIR but failed largely on different questions (Supplementary Section~\ref{sec:S3}), whereas Microsoft-Decision-1 was less accurate than Jev in both benchmarks and, on RheumBench, no more accurate than a 12-billion-parameter generative model run on a laptop.

\subsection{Value of the probabilities}\label{sec:disc-prob}

The feature that distinguishes System One models is the probability attached to each answer. Jev's probabilities on examination questions were close to observed frequencies and ranked its answers well. In a deferral workflow~\citep{Dvijotham2023}, the 0.90 threshold would have left about a quarter of the questions for a clinician to review, with an error rate of 1.5\% among accepted answers. However, clinical thresholds must be chosen and validated for each task, and they must also be set for each model. For instance, at the same threshold, Microsoft-Decision-1's accepted answers contained more errors than Clef's.

Our RheumBench analysis showed that observed calibration differed across tasks. On diagnostic vignettes, Jev's probabilities were overconfident and its accepted answers less accurate; GPT-6 Luna's probabilities, which had been lower than its accuracy on examination questions, had the highest observed correctness AUROC among the System One conditions; and Clef had the lowest observed ECE but was rarely confident enough to reach the threshold. Calibration measured on one task therefore cannot be assumed to transfer to another, even within a specialty, although estimates based on 52 cases are imprecise.

The value of a native probability was clearest in the comparison with the local generative models, whose stated probabilities discriminated poorly between their correct and incorrect answers in both benchmarks, in agreement with reports that confidence verbalised by LLMs is a poor guide to correctness~\citep{Savage2025}. Against GPT-6 Sol, in contrast, calibration gave the System One models no practical advantage, as in the previous benchmark~\citep{MadridGarcia2026jev}. All the frontier model's answers with a stated probability of at least 0.90 were correct, and they covered almost all examination questions and 44\% of the RheumBench cases.

Probabilities over the listed options do not explicitly represent the possibility that the correct answer is absent. When the reference diagnosis was not among the candidates, the 0.90 threshold would still have accepted a forced answer for six of the nine conditions, and only GPT-6 Luna, Microsoft-Decision-1 and Laya in run 1 selected None of the above in those cases. In clinical tasks, the answer options must therefore be designed and validated as carefully as the thresholds, for example with an explicit ``not documented'' or ``other'' answer whose use is evaluated.

\subsection{Language, repeatability and deployment}\label{sec:disc-lang}

We found no evidence that Spanish reduced accuracy in RheumaMIR. Jev was not fully deterministic, and Microsoft-Decision-1 changed answers between runs, whereas GPT-6 Luna and Clef returned identical answers and probabilities in every run. GPT-6 Sol with medium reasoning was not fully repeatable either. It changed 6 of the 52 RheumBench answers, more than any System One model, although its accuracy did not change. Jev's latency was 4--16 times lower than that of GPT-6 Sol and its cost 50--180 times lower, and the other remote System One models were also faster and cheaper than GPT-6 Sol, differences that would matter if every note or referral in a health system were processed~\citep{Klang2024}. Finally, although open models keep patient data within the institution, the local generative models were less accurate, slower and less stable than Jev.

\subsection{Implications}\label{sec:disc-impl}

Examination questions and vignettes with candidate diagnoses are not the intended tasks of System One models, and these results do not show that any of them can classify patients, triage referrals or extract disease activity from notes. They show that the best of them have enough knowledge of the specialty, in Spanish and English, for such applications to be worth testing, and that their probabilities could help decide which answers a rheumatologist reviews. 

Decision and System One models could support rheumatologists in the repeated, clearly defined judgements of routine care. Because these models choose among answers rather than generate them, their natural diagnostic role may be to check candidate diagnoses proposed by clinicians or generative models. They could also act as the decision layer of agentic AI systems, in which a language model decides at each step what should happen next and acts through tools such as the electronic health record or clinical calculators~\citep{MadridGarcia2025agents}.

These uses need their own validation. RheumBench showed that, when the correct option was missing, the models rarely recognised it. Future studies should use real clinical text from local records in the local language, judgements decomposed into atomic questions, clinician-adjudicated references, answer options that allow for missing information, thresholds chosen on separate data for each model and task, and evaluation of the complete model-clinician workflow, to establish whether such workflows reduce workload without increasing errors.

\subsection{Limitations}\label{sec:limitations}

First, neither benchmark represents routine clinical care. Examination questions are written to be unambiguous and always contain the answer, image-dependent questions were excluded, and the RheumBench candidate lists narrow the diagnostic search. Second, the samples were small. Third, training-data contamination cannot be excluded, because the MIR questions are public and 60 of the 70 RheumBench vignettes are published case reports. Fourth, neither language version reproduces the examination wording exactly. The English items were machine translations revised with a language model, and the same model was used to correct spelling, accent, punctuation and transcription errors in the Spanish text. Finally, some endpoints were in public beta or alpha, no analysis plan was registered, and remote services may change without notice; the results describe the versions available between 5 and 10 October 2026.

\section{Conclusion}\label{sec:conclusion}

On rheumatology examination questions in Spanish and English, Jev 1.13 answered more than nine questions in ten correctly; its probabilities identified a large subset of answers containing few errors, and its answers were fast and inexpensive. On diagnostic case vignettes with up to 18 candidate diagnoses, accuracy was lower for every model, Jev's probabilities were overconfident, and all System One models remained below GPT-6 Sol. Microsoft-Decision-1 was less accurate and overconfident in both benchmarks, and the locally run System One model Laya performed at or near chance. Results obtained with one System One model or on one task therefore cannot be generalised to others. These findings support evaluating System One models, with deferral of uncertain answers to the rheumatologist and answer options that allow for missing information, on structured decisions such as applying classification criteria, triaging referrals and extracting information from clinical notes, using real clinical text before any use in patient care.

\section*{Declarations}

\subsection*{Author contributions (CRediT)}

\begin{itemize}
\item \textbf{Alfredo Madrid-García:} Conceptualization; Methodology; Software; Formal analysis; Investigation; Data curation; Visualization; Writing -- original draft; Writing -- review \& editing; Resources; Project administration.

\item \textbf{Diego Benavent:} Data curation; Writing -- review \& editing.

\item \textbf{Fabian Lechner:} Data curation; Resources; Conceptualization

\item \textbf{Phillip Kremer:} Data curation; Resources; Writing -- review \& editing

\item \textbf{Johannes Knitza:} Conceptualization; Resources; Writing -- review \& editing.

\item \textbf{Beatriz Merino-Barbancho:} Conceptualization; Software; Visualization; Writing -- review \& editing.
\end{itemize}

\subsection*{Funding}

This study did not receive any specific grant from funding agencies in the public, commercial or not-for-profit sectors. Johannes Knitza was supported in part by the European Union's Horizon Europe program under grant agreement No. 101080711 (SPIDERR).

\subsection*{Conflicts of interest}

\begin{itemize}
\item \textbf{Alfredo Madrid-García} is employed as a data scientist by Roche. This work was carried out independently of that employment, without the use of company facilities, tools, data or confidential information. The views expressed are those of the author and do not represent the views or official position of Roche. 

\item \textbf{Diego Benavent} has received consultancies/fees from: Abbvie, Bristol-Myers Squibb, Eli Lilly, Janssen, Savana Research (part-time work), UCB. He is also a member of the editorial board of EULAR Rheumatology Open.

\item \textbf{Phillip Kremer} has received payment or honoraria for lectures, presentations or educational events from AlfaSigma, and support for attending meetings and/or travel from AlfaSigma, Boehringer Ingelheim and UCB.

\item \textbf{Johannes Knitza} has received consultancies/fees from Abbvie, AstraZeneca, BMS, Boehringer Ingelheim, Chugai, Fraunhofer, Fachverband Rheumatologische Fachassistenz, GAIA, Galapagos, GSK, Janssen, Lilly, Medac, Novartis, Pfizer, Rheumaakademie, Sanofi, Sobi, UCB and is associate editor of EULAR Rheumatology Open.
\end{itemize}

The other authors declare no competing interests.

\subsection*{Data availability}

\begin{itemize}
\item The expanded RheumaMIR dataset is available from Zenodo~\citep{MadridGarcia2026rheumamir}. The complete requests and responses of all runs, and the notebooks used to run the models and analyse the results, are available as supplementary material (SM\_code\_dataset\_statistics)

\item The RheumBench vignettes belong to the RheumBench study~\citep{Lechner2026}. The RheumBench data, the requests and responses of all runs and the notebooks will be made available by JK upon reasonable request.
\end{itemize}

\subsection*{Supplementary material}

\begin{itemize}
  \item \textbf{SM\_code\_dataset\_statistics:} Datasets, model outputs, and notebooks for model evaluation and statistical analysis on RheumaMIR. \url{https://zenodo.org/records/23741216}
  \item \textbf{SM\_supplementary\_text:} Supplementary methods and results, detailing dataset preparation, model configurations, statistical comparisons, calibration, errors with the highest probabilities, sensitivity analyses, repeatability, response time, and cost. In this preprint, it follows the references (Sections~\ref{sec:S1}--\ref{sec:S8}).
  \item \textbf{SM\_TRIPOD-LLM:} Completed TRIPOD-LLM reporting checklist indicating where each applicable reporting item is addressed in the manuscript and supplementary material. \url{https://zenodo.org/records/23741216}
\end{itemize}

\subsection*{Ethics approval}

RheumaMIR consists of publicly available examination questions published by the Spanish Ministry of Health; no patients were recruited, and no clinical records or personal data were used. RheumBench consists of published case reports and de-identified cases; for the RheumBench study, the Ethics Committee of Philipps-University Marburg confirmed that formal approval was not required (26--266 ANZ)~\citep{Lechner2026}.

\subsection*{Patient consent for publication}

Not applicable.

\subsection*{Generative AI statement}

Generative AI tools (Anthropic Claude Opus 5.5 and OpenAI Codex 6.1 Sol) were used to assist with drafting, coding, and language editing. Claude Opus 5.5 (reasoning setting High) was also used to review the English translations of the RheumaMIR questions and to correct language errors in their Spanish text, with the prompt reported in Supplementary Section~\ref{sec:S1}. All AI-assisted content was reviewed and edited by the authors.

\subsection*{Acknowledgements}

Not applicable.

\bibliographystyle{unsrtnat}
\bibliography{references}

% ======================================================================
% Supplementary Material
% ======================================================================
\clearpage
\setcounter{section}{0}
\setcounter{table}{0}
\setcounter{figure}{0}
\renewcommand{\thesection}{S\arabic{section}}
\renewcommand{\thetable}{S\arabic{table}}
\renewcommand{\thefigure}{S\arabic{figure}}
\renewcommand{\theHsection}{S\arabic{section}}
\renewcommand{\theHtable}{S\arabic{table}}
\renewcommand{\theHfigure}{S\arabic{figure}}

\begin{center}
  {\Large\bfseries Supplementary Material}\\[6pt]
  {\large Decision and System One models in rheumatology: accuracy and calibration on specialty-training examination questions and diagnostic case vignettes}\\[4pt]
  \emph{Alfredo Madrid-Garc\'{\i}a, Diego Benavent, Fabian Lechner, Phillip Kremer, Johannes Knitza, Beatriz Merino-Barbancho}
\end{center}

\paragraph{Contents}\mbox{}
\begin{itemize}
  \item[] \ref{sec:S1}. Datasets and their preparation: \ref{sec:S1.1}. RheumaMIR (Table~\ref{tab:S1}); \ref{sec:S1.2}. RheumBench
  \item[] \ref{sec:S2}. Model configurations and inputs (Table~\ref{tab:S2})
  \item[] \ref{sec:S3}. Statistical methods and paired comparisons (Table~\ref{tab:S3})
  \item[] \ref{sec:S4}. Probability results and sensitivity analyses (Table~\ref{tab:S4}; Figure~\ref{fig:S1})
  \item[] \ref{sec:S5}. Repeatability, latency and cost (Tables~\ref{tab:S5}--\ref{tab:S6})
  \item[] \ref{sec:S6}. RheumBench: reference diagnosis not among the options (Table~\ref{tab:S7})
  \item[] \ref{sec:S7}. RheumaMIR: errors with the highest probabilities (Table~\ref{tab:S8})
  \item[] \ref{sec:S8}. Reproducibility
\end{itemize}

\section{Datasets and their preparation}\label{sec:S1}

Both datasets were converted into JSON files with the same structure: one record per item, holding the text of each language and the official answer as an option number. The models received only the content of these files. The structure and preparation code for RheumaMIR can be found in \emph{SM\_code\_dataset\_statistics.zip.} For RheumBench, it will be made available by JK upon reasonable request.

\subsection{RheumaMIR}\label{sec:S1.1}

\textbf{Source.} RheumaMIR comprises rheumatology-related MIR examination questions from 2009--2010 to 2022--2023, including those from overlapping specialties but excluding image-dependent questions~\citep{MadridGarcia2023,MadridGarcia2023rheumamir}. Questions had five answer options until 2014--2015 and four from 2015--2016 onwards (Table~\ref{tab:S1}). Each record includes the examination year, question type (clinical case or factual), patient sex, disease category, invalidation status, official answer, Spanish text and an English translation initially generated using DeepL. For this study, the dataset was consolidated into a single JSON file retaining each question's pre-revision text. This expanded version is deposited in Zenodo alongside a report comparing each question with its pre-revision version~\citep{MadridGarcia2026rheumamir}. Questions are identified by dataset ID (1--175).

\textbf{Duplicate and invalidated records.} The dataset contains 175 records, 145 records from the 2009--2010 to 2022--2023 examinations and 30 added from the 2023--2024 to 2025--2026. Records 120 and 140 duplicate questions 24 and 73, respectively; they were retained for traceability, marked as excluded and never sent to the models, leaving 173 evaluated questions. Question 111, a five-option clinical case from 2010--2011, is labelled invalidated but retains its legacy key (option 3). It was evaluated but excluded from all performance analyses, leaving 172 questions.

\textbf{Added questions.} A rheumatologist (DB) selected 30 additional questions: 10 from 2023--2024 (MIR 2024), 11 from 2024--2025 (MIR 2025) and 9 from 2025--2026 (MIR 2026). Text was taken from version-0 examination booklets and keys from the Ministry of Health's definitive answer keys. Two image-dependent questions were excluded.

\textbf{Reference-key issues.} The reference answer of every question was the definitive key published by the Ministry of Health. Question 73 was not invalidated. Its excluded duplicate, record 140, carried an erroneous inherited invalidation label, which was corrected in the current dataset and preserved in the historical fields for traceability. Question 73 was included in the primary analysis; record 140 remained excluded as a duplicate. Question 93 (2011--2012, factual, five options) asks which method is most effective for preventing hip fracture (options: oestrogens; calcium and vitamin D; bisphosphonates; physical activity; calcitonin). Its inherited key is option 4, physical activity. Both GPT-6 Sol conditions, Jev, GPT-6 Luna, Microsoft-Decision-1 and Qwen3-8B selected option 3, bisphosphonates, in both languages, and Laya selected it in English (option 2 in Spanish). Clef selected the keyed option in both languages, as did Gemma 4 12B in run 1.

\textbf{Translation and text revision.} English text was revised for 129 of the 143 historical questions to correct mistranslations and omissions, clarify patient sex where stated in Spanish, and standardise terminology. In 10 questions, male wording was inferred from masculine Spanish grammar. Numbers, doses, units, option numbers and order were preserved, as were false statements in options. Spanish spelling, accent, punctuation and transcription errors were corrected (125 logged corrections across 72 questions), and final full stops were added to some options. The instruction ``Justifique su respuesta'' (``Justify your answer''), present in both languages in 144 of 145 historical questions and no new questions, was removed; it was misspelt ``Jusitifique'' in two Spanish stems. Revision in October 2026 included two complete reviews of all translations against the Spanish. The dataset retains every question's pre-revision text, and the deposited comparison report documents all changes. Translations and Spanish text were reviewed using Claude Opus 5.5 (Anthropic; reasoning setting High) with the following verbatim prompt: ``\emph{Review all the questions in Spanish and their English translations. Check that each English translation accurately reflects the original meaning in Spanish, as it was generated automatically. Correct any translation, grammar, spelling, or wording errors, while preserving the original meaning as faithfully as possible. Also review the Spanish questions to identify and correct any typos, spelling mistakes, or grammatical errors}.''. New questions were translated directly into English.

\textbf{Labels.} Question type and disease category were inherited from RheumaMIR and assigned to the added questions with the same criteria. The patient-sex label, not used in the analyses, was relabelled from male to not specified in questions 85 and 95, whose Spanish stems do not state it, and from ``\emph{does not apply}'' to male in added question 172, a clinical case whose sex was inferred from masculine Spanish grammar.

\begin{table}[htbp]
\caption{Characteristics of the evaluated RheumaMIR questions.}\label{tab:S1}
\centering\small
\begin{tabular}{@{}lccc@{}}
\toprule
Characteristic & \makecell[b]{Historical (2009--2010\\to 2022--2023), n=143} & \makecell[b]{New (2023--2024\\to 2025--2026), n=30} & \makecell[b]{Total,\\n=173}\\
\midrule
\multicolumn{4}{@{}l}{\textbf{\emph{Number of options, n (\%)}}}\\
\quad Five (2009--2010 to 2014--2015) & 63 (44.1) & 0 (0.0) & 63 (36.4)\\
\quad Four (2015--2016 onwards) & 80 (55.9) & 30 (100.0) & 110 (63.6)\\
\addlinespace[3pt]
\multicolumn{4}{@{}l}{\textbf{\emph{Type of question, n (\%)}}}\\
\quad Clinical case & 95 (66.4) & 21 (70.0) & 116 (67.1)\\
\quad Factual & 48 (33.6) & 9 (30.0) & 57 (32.9)\\
\addlinespace[3pt]
\multicolumn{4}{@{}l}{\textbf{\emph{Disease addressed, n (\%)}}}\\
\quad Vasculitis & 23 (16.1) & 9 (30.0) & 32 (18.5)\\
\quad Bone metabolism & 20 (14.0) & 2 (6.7) & 22 (12.7)\\
\quad Systemic lupus erythematosus & 16 (11.2) & 3 (10.0) & 19 (11.0)\\
\quad Spondyloarthritis & 14 (9.8) & 4 (13.3) & 18 (10.4)\\
\quad Rheumatoid arthritis & 12 (8.4) & 1 (3.3) & 13 (7.5)\\
\quad Infectious arthritis & 11 (7.7) & 0 (0.0) & 11 (6.4)\\
\quad Microcrystalline arthritis & 9 (6.3) & 2 (6.7) & 11 (6.4)\\
\quad Systemic sclerosis & 8 (5.6) & 1 (3.3) & 9 (5.2)\\
\quad Others & 30 (21.0) & 8 (26.7) & 38 (22.0)\\
\bottomrule
\end{tabular}
\tabnote{%
Counts include invalidated question 111 (historical, five options, clinical case, bone metabolism). The 172 analysed questions comprise 142 historical and 30 new questions, 115 clinical cases and 57 factual questions, and 110 four-option and 62 five-option questions. Disease categories are inherited dataset labels, shown here in current terminology (dataset labels: Spondyloarthropathies, Infective arthritis, Scleroderma and Other); ``Others'' includes amyloidosis, autoinflammatory syndromes, fibromyalgia, IgG4-related disease, inflammatory myopathy, osteoarthritis, sarcoidosis and Sjögren's disease, among others.
}
\end{table}

\subsection{RheumBench}\label{sec:S1.2}

RheumBench comprises 100 cases, 70 diagnostic and 30 management, compiled by board-certified rheumatologists~\citep{Lechner2026}. The diagnostic cases include 60 published case reports of rare rheumatic diseases and clinically important differential diagnoses, curated previously~\citep{Kremer2025}, and 10 de-identified cases from University Hospital Hamburg, added to reduce training-data contamination risk. Two rheumatologists independently reviewed each case for completeness and quality and assessed its reference diagnosis for clinical plausibility. Each case includes patient age, sex, symptoms and available laboratory, examination and imaging findings. In the original benchmark, systems provided free-text answers, scored by a panel of LLM judges against case-specific rubrics. This study used only the 70 diagnostic cases, excluding management cases and rubrics and all collective diseases suggestions from the previously investigated LLMs. For 18/70 cases no correct diseases suggestion was initially proposed by the LLMs. 

The evaluation notebook generated two option sets. The listed set (primary) retains each case's k candidate diagnoses, including cases without a correct option (n=18). The None of the above set (secondary) adds ``\emph{None of the above}'' as option k + 1 to every case. This is correct for those 18 cases; answers for the remaining 52 are unchanged, giving all 70 cases a correct option.

\section{Model configurations and inputs}\label{sec:S2}

Table~\ref{tab:S2} lists the evaluated models, their access routes and settings, and the considerations relevant to the interpretation of their results.

\begin{table}[htbp]
\caption{Evaluated models and configurations: version, access, settings and considerations for interpretation.}\label{tab:S2}
\centering\small
\setlength{\tabcolsep}{4pt}
\begin{tabularx}{\linewidth}{@{}>{\raggedright\arraybackslash\hangindent=1em\hangafter=1}p{2.9cm}>{\raggedright\arraybackslash}p{4.1cm}LL@{}}
\toprule
Model and version (developer) & Access and identifier & Settings & Considerations\\
\midrule
\multicolumn{4}{@{}l}{\textbf{\emph{System One models}}}\\
\quad Jev 1.13 (TypeSafe AI) & Remote, OpenRouter (typesafe/jev-1.13); returned as typesafe/jev-1.13-20260917 & Single choice question; native probabilities & Architecture undisclosed; English described by the developer as its primary training language\\
\quad GPT-6 Luna (OpenAI) & Remote, OpenAI Decisions API (gpt-6-luna; public beta); no dated snapshot & Single choice question without fallback choice; native probabilities & Not described by OpenAI as a System One model; architecture undisclosed; returns the selected option and a probability for each option without generating text\\
\quad Clef (Cloudflare)~\citep{Chen2026} & Remote, OpenRouter Decisions endpoint (cloudflare/clef; alpha); provider pinned to Cloudflare Workers AI, fallbacks disabled; no dated snapshot & Request of Jev plus provider field; native probabilities & Open weights; 27 billion parameters; fine-tuned from Qwen3.8-27B\\
\quad Microsoft-Decision-1 (Microsoft) & Remote, OpenRouter Decisions endpoint (microsoft/microsoft-decision-1; alpha), returned as microsoft/microsoft-decision-1-20261009; provider pinned to Azure, fallbacks disabled & Request of Jev plus provider field; native probabilities & Text only; post-trained from Qwen3.5-9B; weights not public and updated continually, so the returned identifier names no fixed snapshot\\
\quad Laya 0.3.20 (Convai Innovations)~\citep{ConvaiInnovations2026} & Local (Apple GPU); English checkpoint (ModernBERT-large, 421 million parameters) for English RheumaMIR items and RheumBench, multilingual checkpoint (mmBERT-base) for Spanish items & Language passed explicitly; default input limits; no recalibration & Open source; shortened option text in two RheumaMIR questions (three question--language pairs) and, with a 512-token limit, cut the end of 10 RheumBench vignettes (9 of the 52 analysed)\\
\addlinespace[3pt]
\multicolumn{4}{@{}l}{\textbf{\emph{Primary comparator}}}\\
\quad GPT-6 Sol, medium reasoning (OpenAI) & Remote, OpenAI chat completions API (gpt-6-sol); no dated snapshot & Medium reasoning effort; strict JSON schema & Frontier generative model; output constrained by a strict JSON schema to one option and a probability for each option\\
\addlinespace[3pt]
\multicolumn{4}{@{}l}{\textbf{\emph{Secondary comparators}}}\\
\quad GPT-6 Sol, no reasoning (OpenAI) & Remote, OpenAI chat completions API (gpt-6-sol); no dated snapshot & No reasoning; strict JSON schema & As above, without reasoning\\
\quad Gemma 4 12B (Google DeepMind)~\citep{Farabet2026,GoogleDeepMind2026gemma} & Local; mlx-community/gemma-4-12B-it-4bit (4-bit MLX conversion), MLX-LM server~\citep{MLXLM} & Thinking disabled; temperature 1.0, top-p 0.95, top-k 64, as recommended in the model card; up to 2,048 output tokens & Asked to answer only with a JSON object\\
\quad Qwen3-8B (Qwen Team)~\citep{QwenTeam2025qwen3} & Local; mlx-community/Qwen3-8B-4bit (4-bit MLX conversion), MLX-LM server & Thinking disabled; temperature 0.7, top-p 0.8, top-k 20, as recommended in the model card; up to 2,048 output tokens & Asked to answer only with a JSON object\\
\bottomrule
\end{tabularx}
\tabnote{%
Remote models were accessed through the application programming interfaces named; local models ran on a MacBook Pro (Apple M1 Pro, 16 GB). In RheumaMIR, nine conditions were evaluated in two languages and three runs: 54 experiments and 9,342 saved evaluations, including invalidated question 111. RheumaMIR was run between 5 and 10 October 2026. In RheumBench, the same nine conditions were evaluated with two option sets and three runs: 54 experiments and 3,780 saved evaluations. The listed set was run on 8 October 2026 and the None of the above set on 9 October 2026; Microsoft-Decision-1 answered both sets on 10 October 2026. Thinking-enabled conditions of the local generative models were not run. No model was fine-tuned or recalibrated, and no benchmark-specific parameter search was performed.
}
\end{table}

\textbf{Hardware and software.} Local models ran on an Apple M1 Pro MacBook Pro with 16 GB of unified memory, with Python 3.13, Laya 0.3.20 and MLX-LM 0.32.0, whose server ran with the prompt cache disabled (\texttt{-{}-prompt-cache-size 0}), so that every request processed its full prompt.

\subsection{System One requests}\label{sec:S2.1}

\textbf{RheumaMIR.} The instruction was ``Which option is the correct answer to this question?'' in English and ``¿Cuál es la respuesta correcta a esta pregunta?'' in Spanish. Jev received each item as one request to the OpenRouter System One endpoint, with the stem as state and one choice question:

\begin{Verbatim}[fontsize=\small,frame=leftline,xleftmargin=1em,framesep=2mm,samepage=true]
{"model": "typesafe/jev-1.13", "state": "{stem}",
 "questions": {"answer": {"type": "choice",
   "instructions": "{instruction}",
   "criteria": {"1": "{option 1}", "2": "{option 2}", "3": "{option 3}", "4": "{option 4}"}}}}
\end{Verbatim}

Laya received the same state and question through its local package, with the language passed explicitly; with its default input limits, it shortened the options of two questions (three question--language pairs), and no stem or instruction was cut.

GPT-6 Luna received the same question through the OpenAI Decisions API (public beta), with the stem as input and one choice per option, without a fallback choice. Clef and Microsoft-Decision-1 received the Jev request unchanged apart from the model identifier and a field pinning the provider (Cloudflare or Azure) through OpenRouter's Decisions endpoint (alpha); every answer was served by the requested provider, and the longest input was well within the context limit of each model.

\textbf{RheumBench.} The requests had the same structure, with the instruction ``Which option is the most likely diagnosis?'', the vignette as state (for GPT-6 Luna, as input) and the k candidate diagnoses, numbered 1 to k, as criteria (for GPT-6 Luna, as choices); in the None of the above set, option k + 1 was added. Laya's English checkpoint shortened no option or instruction but cut the end of 10 vignettes (9 of the 52 analysed).

\subsection{Generative prompts and response validation}\label{sec:S2.2}

\textbf{RheumaMIR.} GPT-6 Sol received a single user message without a system prompt. English template (five-option questions add option 5 and refer to five probabilities):

\begin{Verbatim}[fontsize=\small,frame=leftline,xleftmargin=1em,framesep=2mm,samepage=true]
Which option is the correct answer to this question?

Question:
{stem}

Options:
1. {option 1}
2. {option 2}
3. {option 3}
4. {option 4}

Return the number of the correct answer in "answer" and, in "probabilities", your probability
(between 0 and 1) that each option is the correct answer. The four probabilities must sum to 1.
\end{Verbatim}

Spanish template:

\begin{Verbatim}[fontsize=\small,frame=leftline,xleftmargin=1em,framesep=2mm,samepage=true]
¿Cuál es la respuesta correcta a esta pregunta?

Pregunta:
{stem}

Opciones:
1. {option 1}
2. {option 2}
3. {option 3}
4. {option 4}

Devuelve el número de la respuesta correcta en "answer" y, en "probabilities", tu probabilidad
(entre 0 y 1) de que cada opción sea la respuesta correcta. Las cuatro probabilidades deben
sumar 1.
\end{Verbatim}

For RheumBench, GPT-6 Sol received the following template (k is the number of options; with None of the above, it is the last option); the local generative models received the same message followed by the JSON format line described below:

\begin{Verbatim}[fontsize=\small,frame=leftline,xleftmargin=1em,framesep=2mm,samepage=true]
Which option is the most likely diagnosis?

Case:
{vignette}

Options:
1. {option 1}
2. {option 2}
...
k. {option k}

Return the number of the most likely diagnosis in "answer" and, in "probabilities", your
probability (between 0 and 1) that each option is the correct diagnosis. The k probabilities
must sum to 1.
\end{Verbatim}

The strict JSON schema required an answer string from the available option numbers and a numerical probability for every available option. The local generative models received the same message followed by ``Answer only with a JSON object in this format, with no other text:'' (Spanish: ``Responde únicamente con un objeto JSON con este formato, sin ningún otro texto:'') and a JSON example with the answer and all probability keys, because their server cannot enforce a schema.

\textbf{Retries and output limits.} Unsuccessful requests were attempted up to three times. Every request ultimately returned a valid answer, 23 after a second attempt (17 in RheumaMIR and 6 in RheumBench). For the local generative models, the parser read the first JSON object in each response, after removing any thinking text (present in four Gemma 4 12B RheumaMIR responses, two of them in run 1, although thinking was disabled; removing it did not change their answers), and required an available answer and finite, non-negative probabilities with a positive total. Probabilities written for options that had not been offered (nine Gemma 4 12B responses, two in RheumaMIR and seven in RheumBench; non-zero, 0.01, in two RheumBench responses from runs 2 and 3 of cases without the reference diagnosis) were discarded before rescaling; the returned answer was not affected, and no run-1 estimate changed.

\textbf{Probability vectors.} Native probabilities were analysed as returned; Jev's are rounded to two decimals, so 6 of its 1,038 RheumaMIR vectors summed to 0.99. Stated vectors that did not sum to 1 ($\pm$0.01) occurred in 43 of 1,038 Gemma 4 12B and 23 of 1,038 Qwen3-8B responses; all stated vectors were rescaled to sum to 1 over the offered options. The selected option had a lower probability than another option in one Jev response and 80 Qwen3-8B responses across the three runs, and the selected option shared the highest probability with another option in three Microsoft-Decision-1, two Jev, four Gemma 4 12B and 38 Qwen3-8B responses; the returned answer was always scored.

In RheumBench (listed option set; 210 responses per model, 70 cases in three runs), all native probability vectors returned by the System One models and GPT-6 Luna summed to 1 within 0.01 (Jev 0.99--1.00). Stated vectors that did not sum to 1 ($\pm$0.01) occurred in 120 of 210 Gemma 4 12B responses, 164 of 210 Qwen3-8B responses, 4 of 210 GPT-6 Sol responses without reasoning and none with medium reasoning; they were rescaled as above. The selected option had a lower probability than another option in one Jev response and two Qwen3-8B responses, and shared the highest probability with another option in two Microsoft-Decision-1 responses, one Gemma 4 12B response and two Qwen3-8B responses; the returned answer was always scored.

\section{Statistical methods and paired comparisons}\label{sec:S3}

Wilson intervals describe single proportions. Exact McNemar tests use only the questions answered correctly by one of the two conditions being compared. Two thousand bootstrap resamples of question identifiers, generated with NumPy seed 2026, provide percentile intervals for paired differences and AUROC; each resample keeps languages and models paired. Holm adjustment was applied to the ten comparisons of System One models with GPT-6 Sol (medium) and, separately, to the five System One language comparisons. The secondary comparators were described without significance tests.

In RheumBench, the same methods were applied to the 52 cases with the reference diagnosis among the options, with bootstrap resamples of cases. Holm adjustment was applied to the five comparisons of System One models with GPT-6 Sol (medium) for accuracy.

\textbf{Errors in RheumaMIR.} GPT-6 Sol with medium reasoning answered 11 of Jev's 12 Spanish errors and 13 of its 14 English errors correctly, and all of Clef's errors (Table~\ref{tab:S3}). Jev shared 4 of Clef's 14 Spanish errors and 5 of its 12 English errors, so the two models failed largely on different questions, and GPT-6 Luna's three errors in each language other than question 93 differed from Jev's. GPT-6 Sol's only discordance with the answer key in run 1, in both conditions and languages, was question 93 (2011--2012) on hip-fracture prevention: GPT-6 Sol, Jev, GPT-6 Luna, Microsoft-Decision-1 and Qwen3-8B selected bisphosphonates (Laya in English only), whereas Clef selected the keyed answer, physical activity (Section~\ref{sec:S1.1}); in run 3, GPT-6 Sol without reasoning also missed question 14 in Spanish. A negatively worded question on temporal arteritis was a confident error (selected-answer probability 0.90--0.98) of Jev, Clef and Microsoft-Decision-1 in both languages (question 33; Table~\ref{tab:S8}); with question 93, it accounted for the two errors among Jev's answers retained at the 0.90 threshold in each language.

\begin{table}[htbp]
\caption{Paired comparisons of the System One models with GPT-6 Sol (medium reasoning).}\label{tab:S3}
\centering\small
\begin{tabular}{@{}lcccc@{}}
\toprule
System One model & \makecell[b]{Difference,\\pp (95\% CI)} & \makecell[b]{Correct only\\GPT-6 Sol, n} & \makecell[b]{Correct only\\System One model, n} & \makecell[b]{P\\(Holm)}\\
\midrule
\multicolumn{5}{@{}l}{\textbf{\emph{RheumaMIR, Spanish}}}\\
\quad Jev 1.13 & $-$6.4 ($-$10.5 to $-$2.9) & 11 & 0 & 0.005\\
\quad GPT-6 Luna (Decisions API) & $-$1.7 ($-$4.1 to 0.0) & 3 & 0 & 0.500\\
\quad Clef & $-$7.6 ($-$11.6 to $-$3.5) & 14 & 1 & 0.005\\
\quad Microsoft-Decision-1 & $-$11.6 ($-$16.3 to $-$7.0) & 20 & 0 & \textless{}0.001\\
\quad Laya 0.3.20 & $-$77.3 ($-$83.1 to $-$70.9) & 133 & 0 & \textless{}0.001\\
\addlinespace[3pt]
\multicolumn{5}{@{}l}{\textbf{\emph{RheumaMIR, English}}}\\
\quad Jev 1.13 & $-$7.6 ($-$11.6 to $-$3.5) & 13 & 0 & 0.001\\
\quad GPT-6 Luna (Decisions API) & $-$1.7 ($-$4.1 to 0.0) & 3 & 0 & 0.500\\
\quad Clef & $-$6.4 ($-$10.5 to $-$2.3) & 12 & 1 & 0.010\\
\quad Microsoft-Decision-1 & $-$13.4 ($-$18.6 to $-$8.1) & 23 & 0 & \textless{}0.001\\
\quad Laya 0.3.20 & $-$74.4 ($-$80.8 to $-$68.6) & 128 & 0 & \textless{}0.001\\
\addlinespace[3pt]
\multicolumn{5}{@{}l}{\textbf{\emph{RheumBench}}}\\
\quad Jev 1.13 & $-$26.9 ($-$42.3 to $-$9.6) & 18 & 4 & 0.013\\
\quad GPT-6 Luna (Decisions API) & $-$13.5 ($-$25.0 to $-$3.8) & 8 & 1 & 0.045\\
\quad Clef & $-$17.3 ($-$30.8 to $-$3.8) & 11 & 2 & 0.045\\
\quad Microsoft-Decision-1 & $-$36.5 ($-$51.9 to $-$23.1) & 20 & 1 & \textless{}0.001\\
\quad Laya 0.3.20 & $-$67.3 ($-$78.8 to $-$53.8) & 35 & 0 & \textless{}0.001\\
\bottomrule
\end{tabular}
\tabnote{%
Differences are System One model minus GPT-6 Sol (medium reasoning), in percentage points (pp), with 95\% CIs from 2,000 bootstrap resamples of questions or cases; exact McNemar tests, Holm-adjusted within each family (ten RheumaMIR and five RheumBench comparisons). With few discordant items, a significant comparison can rest on a handful of questions or cases.
}
\end{table}

\section{Probability results and sensitivity analyses}\label{sec:S4}

\begin{figure}[htbp]
\centering
\begin{minipage}[t]{0.49\linewidth}\centering
{\small\textbf{(A)}}\\[2pt]
\includegraphics[width=\linewidth]{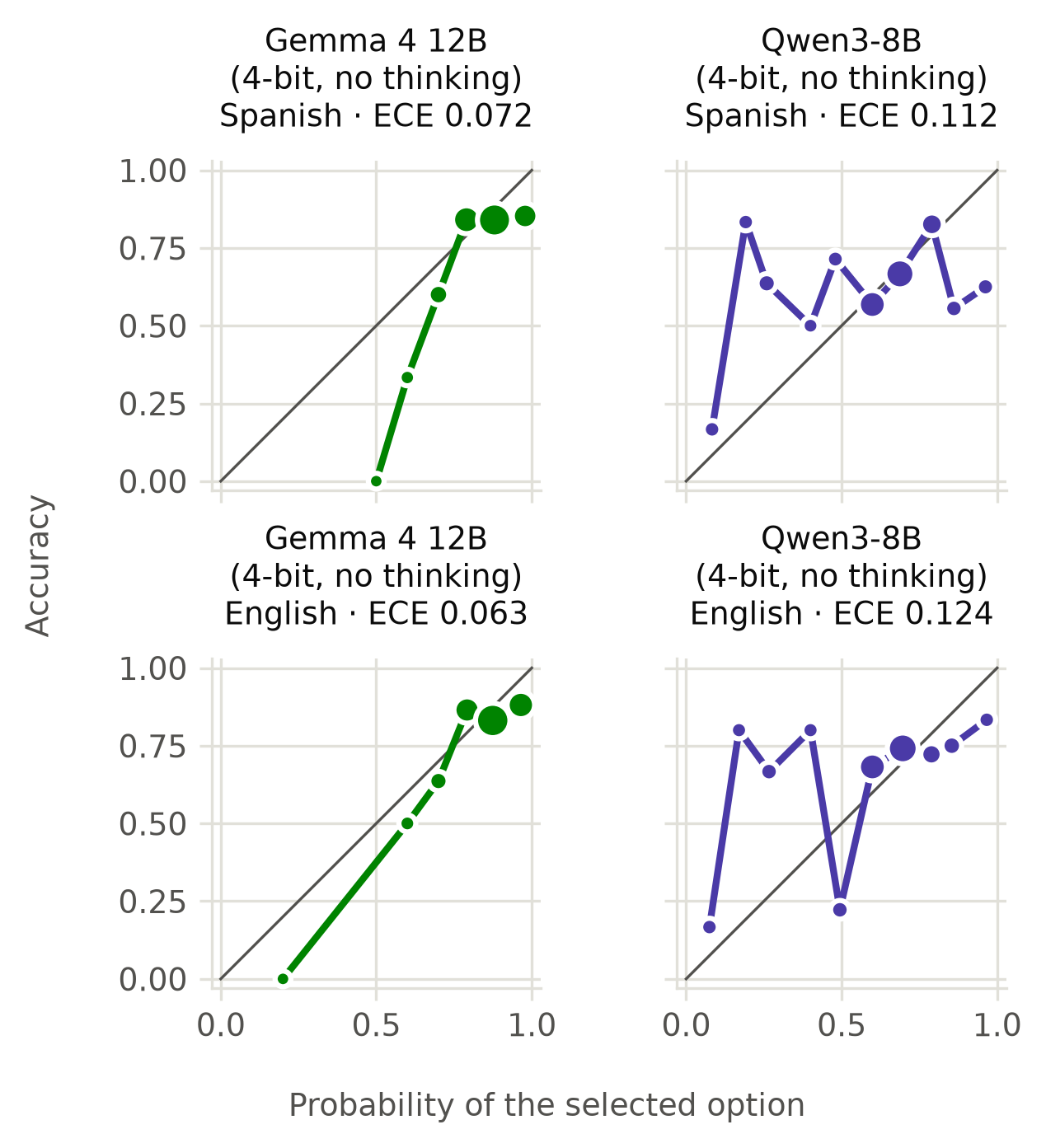}
\end{minipage}\hfill
\begin{minipage}[t]{0.49\linewidth}\centering
{\small\textbf{(B)}}\\[2pt]
\includegraphics[width=\linewidth]{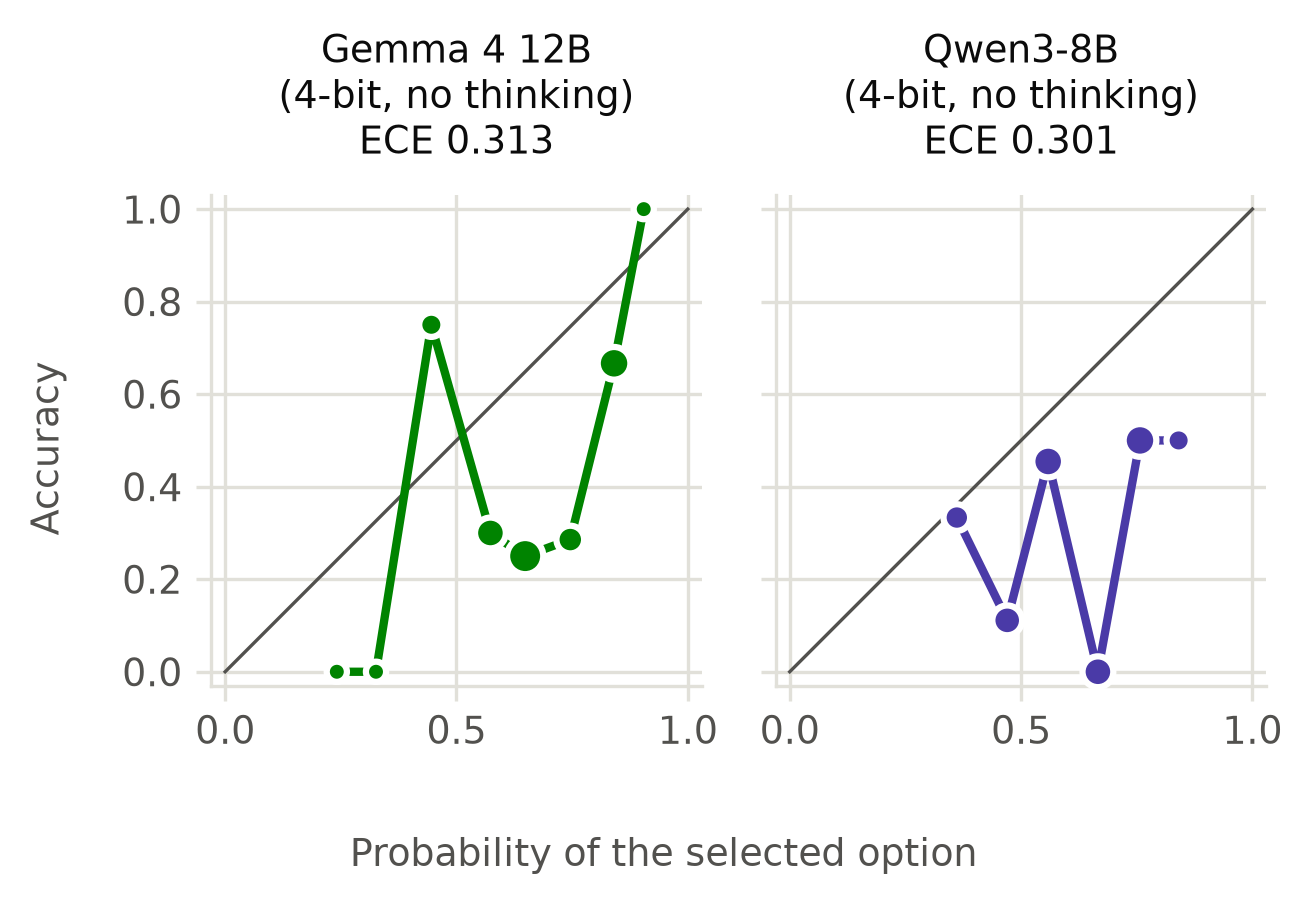}
\end{minipage}
\caption{Reliability diagrams for local generative models with thinking disabled (run 1). (A) RheumaMIR in Spanish (top) and English (bottom). (B) RheumBench, listed option set (52 cases). Each point represents one of ten equal-width bins (closed on the right; Section~\ref{sec:S4}), plotting mean selected-option probability against accuracy. Point size reflects bin count. Points below the diagonal indicate overconfidence; small bins warrant caution. System One models and GPT-6 Sol appear in Figure~\ref{fig:2} of the main text.}\label{fig:S1}
\end{figure}

Probability analyses use selected-answer probabilities, rather than the separate confidence values returned by Jev, GPT-6 Luna, Clef and Microsoft-Decision-1. The over/underconfidence index is the mean selected-answer probability minus accuracy; positive values indicate overconfidence. Expected calibration error (ECE) is the weighted mean absolute difference between mean probability and accuracy across equal-width bins closed on the right, with a probability of 0 in the first bin; ten-bin estimates are descriptive and bin-dependent. The convention matters because Jev, GPT-6 Luna, GPT-6 Sol and the local generative models often give probabilities that lie exactly on a bin edge. Correctness AUROC measures whether correct answers receive higher selected-answer probabilities than incorrect answers and is reported only when at least five answers are correct and five incorrect. Correctness AUROC values are given in Table~\ref{tab:S4}, and reliability diagrams for the local generative models in Figure~\ref{fig:S1}.

\textbf{Sensitivity analyses.} Scoring the highest-probability option instead of the returned answer (retaining the latter when tied for highest probability) changed accuracy by at most one question or case in every condition except Qwen3-8B on RheumaMIR. Its returned answer differed from its highest-probability option in 80 of 1,038 responses; accuracy fell from 64.0\% to 59.3\% in Spanish and from 68.0\% to 65.1\% in English. Laya truncated 9 of the 52 analysed RheumBench vignettes, answering none correctly (random guessing: 9.0\%). It correctly answered 5 of the 43 complete vignettes (11.6\%, 95\% CI 5.1--24.5; random guessing: 9.8\%).

\textbf{Additional probability results.} In RheumaMIR, 10 of Jev's 12 Spanish errors and 12 of its 14 English errors had selected-answer probabilities below 0.9; retained errors had probabilities of 0.90--0.96 (Section~\ref{sec:S3}). GPT-6 Luna's probabilities underestimated accuracy (ECE 0.049 and 0.062), so the 0.90 threshold deferred 37 Spanish and 45 English answers, including only three and four errors, respectively. Four errors per language were insufficient to assess discrimination. GPT-6 Sol with medium reasoning assigned its sole discordant answer a probability below 0.9. At this threshold, Clef retained 129 Spanish (75.0\%) and 137 English (79.7\%) answers; Microsoft-Decision-1 retained 150 (87.2\%) and 138 (80.2\%), respectively (ECE 0.088 and 0.073); and Gemma 4 12B retained 81 (47.1\%) and 83 (48.3\%), with accuracies of 85.2\% and 84.3\% (Table~\ref{tab:2} of the main text). In RheumBench, ECE was 0.113 for GPT-6 Luna and 0.074 for GPT-6 Sol with medium reasoning. Microsoft-Decision-1 discriminated correct from incorrect answers (AUROC 0.811) despite marked overconfidence (ECE 0.331). Laya's mean selected-answer probability was 0.716, compared with 9.6\% accuracy (ECE 0.646). Qwen3-8B's stated probabilities did not always agree with its selected answer (Section~\ref{sec:S2}).

\begin{table}[htbp]
\caption{Correctness AUROC of the selected-answer probability (95\% CI).}\label{tab:S4}
\centering\small
\begin{tabular}{@{}lccc@{}}
\toprule
Model & RheumaMIR, Spanish & RheumaMIR, English & RheumBench\\
\midrule
\multicolumn{4}{@{}l}{\textbf{\emph{System One models}}}\\
\quad Jev 1.13 & 0.904 (0.843--0.956) & 0.916 (0.859--0.964) & 0.737 (0.598--0.859)\\
\quad GPT-6 Luna (Decisions API) & NE & NE & 0.876 (0.773--0.959)\\
\quad Clef & 0.851 (0.754--0.932) & 0.789 (0.641--0.915) & 0.800 (0.668--0.899)\\
\quad Microsoft-Decision-1 & 0.831 (0.752--0.901) & 0.933 (0.891--0.966) & 0.811 (0.679--0.922)\\
\quad Laya 0.3.20 & 0.613 (0.503--0.710) & 0.522 (0.425--0.624) & 0.323 (0.031--0.729)\\
\addlinespace[3pt]
\multicolumn{4}{@{}l}{\textbf{\emph{Primary comparator}}}\\
\quad GPT-6 Sol (medium) & NE & NE & 0.872 (0.749--0.957)\\
\addlinespace[3pt]
\multicolumn{4}{@{}l}{\textbf{\emph{Secondary comparators}}}\\
\quad GPT-6 Sol (none) & NE & NE & 0.782 (0.638--0.909)\\
\quad Gemma 4 12B (4-bit, no thinking) & 0.610 (0.496--0.722) & 0.587 (0.467--0.702) & 0.620 (0.458--0.786)\\
\quad Qwen3-8B (4-bit, no thinking) & 0.574 (0.486--0.660) & 0.605 (0.507--0.692) & 0.570 (0.392--0.745)\\
\bottomrule
\end{tabular}
\tabnote{%
Run 1; RheumBench, listed option set, 52 cases. Correctness AUROC: probability that a correct answer receives a higher selected-answer probability than an incorrect one (0.5, no separation); it is not a diagnostic AUROC. 95\% CIs from 2,000 bootstrap resamples of questions or cases. NE, not estimable (fewer than five correct or five incorrect answers); the near-ceiling accuracy of GPT-6 Sol and GPT-6 Luna in RheumaMIR leaves too few errors to assess their discrimination. Laya's package warned that its RheumBench probabilities for questions with 11 or more options (29 of the 52 cases) were uncalibrated.
}
\end{table}

\section{Repeatability, latency and cost}\label{sec:S5}

In RheumaMIR, GPT-6 Luna, Clef, Laya and GPT-6 Sol with medium reasoning gave the same answer to every analysed question in all three runs (GPT-6 Sol with medium reasoning changed only its answer to invalidated question 111), and GPT-6 Luna and Clef also returned identical probabilities; Jev changed its answer in five question--language pairs (three Spanish, two English) and Microsoft-Decision-1 in six (four Spanish, two English). In RheumBench, Jev changed its answer in 2 of the 52 cases, Microsoft-Decision-1 in 3 within two minutes, GPT-6 Sol in 4 (no reasoning) and 6 (medium reasoning), Qwen3-8B in 5 and Gemma 4 12B in 26 (Table~\ref{tab:S5}); over all 70 cases, including the 18 without the reference diagnosis, the corresponding numbers were 4, 4, 5, 12, 10 and 37. Jev's median latency was four to six times lower than that of GPT-6 Sol in RheumaMIR and 16 times lower than that of GPT-6 Sol with medium reasoning in RheumBench, and its cost 48--77 and 181 times lower. Locally, Laya answered in 0.04--0.11 s, Qwen3-8B in 3.1--7.0 s and Gemma 4 12B in 5.7--39.9 s (Table~\ref{tab:S6}).

\begin{table}[htbp]
\caption{Repeatability over three runs.}\label{tab:S5}
\centering\small
\setlength{\tabcolsep}{4pt}
\panel{(A) RheumaMIR (172 questions per language)}
\begin{tabular}{@{}lcccc@{}}
\toprule
Model & \makecell[b]{Spanish: same answer\\in 3 runs, \%} & \makecell[b]{Spanish: single-run\\accuracy, \%} & \makecell[b]{English: same answer\\in 3 runs, \%} & \makecell[b]{English: single-run\\accuracy, \%}\\
\midrule
\multicolumn{5}{@{}l}{\textbf{\emph{System One models}}}\\
\quad Jev 1.13 & 98.3 & 93.0--93.6 & 98.8 & 91.9--93.0\\
\quad GPT-6 Luna (Decisions API) & 100.0 & 97.7--97.7 & 100.0 & 97.7--97.7\\
\quad Clef & 100.0 & 91.9--91.9 & 100.0 & 93.0--93.0\\
\quad Microsoft-Decision-1 & 97.7 & 86.6--87.8 & 98.8 & 86.0--86.6\\
\quad Laya 0.3.20 & 100.0 & 22.1--22.1 & 100.0 & 25.0--25.0\\
\addlinespace[3pt]
\multicolumn{5}{@{}l}{\textbf{\emph{Primary comparator}}}\\
\quad GPT-6 Sol (medium) & 100.0 & 99.4--99.4 & 100.0 & 99.4--99.4\\
\addlinespace[3pt]
\multicolumn{5}{@{}l}{\textbf{\emph{Secondary comparators}}}\\
\quad GPT-6 Sol (none) & 99.4 & 98.8--99.4 & 100.0 & 99.4--99.4\\
\quad Gemma 4 12B (4-bit, no thinking) & 90.7 & 80.8--81.4 & 85.5 & 79.7--82.6\\
\quad Qwen3-8B (4-bit, no thinking) & 95.9 & 64.0--65.1 & 95.3 & 66.9--68.6\\
\bottomrule
\end{tabular}
\panel{(B) RheumBench (52 cases, listed option set)}
\begin{tabular}{@{}lcc@{}}
\toprule
Model & \makecell[b]{Same answer in\\3 runs, n (\%)} & \makecell[b]{Single-run\\accuracy, \%}\\
\midrule
\multicolumn{3}{@{}l}{\textbf{\emph{System One models}}}\\
\quad Jev 1.13 & 50 (96.2) & 48.1--50.0\\
\quad GPT-6 Luna (Decisions API) & 52 (100.0) & 63.5--63.5\\
\quad Clef & 52 (100.0) & 59.6--59.6\\
\quad Microsoft-Decision-1 & 49 (94.2) & 38.5--40.4\\
\quad Laya 0.3.20 & 52 (100.0) & 9.6--9.6\\
\addlinespace[3pt]
\multicolumn{3}{@{}l}{\textbf{\emph{Primary comparator}}}\\
\quad GPT-6 Sol (medium) & 46 (88.5) & 76.9--76.9\\
\addlinespace[3pt]
\multicolumn{3}{@{}l}{\textbf{\emph{Secondary comparators}}}\\
\quad GPT-6 Sol (none) & 48 (92.3) & 75.0--76.9\\
\quad Gemma 4 12B (4-bit, no thinking) & 26 (50.0) & 38.5--44.2\\
\quad Qwen3-8B (4-bit, no thinking) & 47 (90.4) & 30.8--30.8\\
\bottomrule
\end{tabular}
\tabnote{%
Three runs with identical requests; single-run accuracy is given as the range over the three runs. Identical answers measure repeatability, not correctness: Laya was fully repeatable despite low accuracy. GPT-6 Luna and Clef returned identical answers and probabilities in every run. Microsoft-Decision-1 changed its answer to six RheumaMIR question--language pairs within ten minutes and to three RheumBench cases within two minutes, all under the same returned model identifier. Gemma 4 12B samples at temperature 1.0.
}
\end{table}

\begin{table}[htbp]
\caption{Response time and cost.}\label{tab:S6}
\centering\small
\setlength{\tabcolsep}{4pt}
\panel{(A) RheumaMIR (173 evaluated questions per language)}
\begin{tabular}{@{}lcccc@{}}
\toprule
Model & \makecell[b]{Spanish: latency, s,\\median (IQR)} & \makecell[b]{Spanish: cost per\\1,000 questions, US\$} & \makecell[b]{English: latency, s,\\median (IQR)} & \makecell[b]{English: cost per\\1,000 questions, US\$}\\
\midrule
\multicolumn{5}{@{}l}{\textbf{\emph{System One models}}}\\
\quad Jev 1.13 & 0.33 (0.31--0.36) & 0.021 & 0.33 (0.31--0.37) & 0.020\\
\quad GPT-6 Luna (Decisions API) & 0.21 (0.20--0.24) & 0.027 & 0.21 (0.20--0.23) & 0.025\\
\quad Clef & 0.47 (0.38--0.54) & 0.075 & 0.47 (0.39--0.56) & 0.069\\
\quad Microsoft-Decision-1 & 0.40 (0.38--0.43) & 0.008 & 0.39 (0.37--0.42) & 0.007\\
\quad Laya 0.3.20 & 0.04 (0.03--0.04) & No charge & 0.07 (0.05--0.07) & No charge\\
\addlinespace[3pt]
\multicolumn{5}{@{}l}{\textbf{\emph{Primary comparator}}}\\
\quad GPT-6 Sol (medium) & 2.05 (1.80--2.42) & 1.589 & 2.04 (1.80--2.35) & 1.520\\
\addlinespace[3pt]
\multicolumn{5}{@{}l}{\textbf{\emph{Secondary comparators}}}\\
\quad GPT-6 Sol (none) & 1.36 (1.25--1.53) & 1.022 & 1.39 (1.26--1.53) & 0.963\\
\quad Gemma 4 12B (4-bit, no thinking) & 5.67 (5.21--6.45) & No charge & 20.13 (6.65--27.90) & No charge\\
\quad Qwen3-8B (4-bit, no thinking) & 3.33 (3.06--3.67) & No charge & 3.09 (2.85--3.35) & No charge\\
\bottomrule
\end{tabular}
\panel{(B) RheumBench (70 evaluated cases, listed option set)}
\begin{tabular}{@{}lcc@{}}
\toprule
Model & \makecell[b]{Latency, s,\\median (IQR)} & \makecell[b]{Cost per 1,000\\cases, US\$}\\
\midrule
\multicolumn{3}{@{}l}{\textbf{\emph{System One models}}}\\
\quad Jev 1.13 & 0.41 (0.38--0.44) & 0.031\\
\quad GPT-6 Luna (Decisions API) & 0.21 (0.20--0.23) & 0.045\\
\quad Clef & 0.48 (0.37--0.56) & 0.138\\
\quad Microsoft-Decision-1 & 0.40 (0.38--0.43) & 0.015\\
\quad Laya 0.3.20 & 0.11 (0.08--0.14) & No charge\\
\addlinespace[3pt]
\multicolumn{3}{@{}l}{\textbf{\emph{Primary comparator}}}\\
\quad GPT-6 Sol (medium) & 6.53 (4.37--9.21) & 5.543\\
\addlinespace[3pt]
\multicolumn{3}{@{}l}{\textbf{\emph{Secondary comparators}}}\\
\quad GPT-6 Sol (none) & 2.11 (1.86--2.29) & 1.911\\
\quad Gemma 4 12B (4-bit, no thinking) & 39.86 (28.36--60.32) & No charge\\
\quad Qwen3-8B (4-bit, no thinking) & 7.01 (5.62--8.19) & No charge\\
\bottomrule
\end{tabular}
\tabnote{%
Latency: successful first-attempt requests across the three runs, excluding warm-up calls; RheumBench, listed option set. Latency and cost were computed over all evaluated items, including invalidated RheumaMIR question 111 and the 18 RheumBench cases whose reference diagnosis is not among the options, so they refer to 173 questions per language and 70 cases, not to the 172 questions and 52 cases of the performance analyses. Costs of Jev, Clef and Microsoft-Decision-1 are the charges recorded by OpenRouter; costs of GPT-6 Luna and GPT-6 Sol were estimated from token counts at list prices (GPT-6 Sol: US\$2.00 per million input tokens and US\$10.00 per million output tokens, reasoning tokens included, with no cached input; GPT-6 Luna: US\$0.10 per million input tokens); all costs are those of the saved attempt of each request, so the failed first attempts of four retried requests to paid services, and connection tests, are not included. Local models ran on a MacBook Pro (Apple M1 Pro, 16 GB) and have no per-request charge, which does not mean zero hardware or energy cost. Remote latencies include the network and local latencies depend on the laptop, so timings are not a matched-hardware comparison.
}
\end{table}

\section{RheumBench: reference diagnosis not among the options}\label{sec:S6}

In the 18 cases whose reference diagnosis was absent from the options, no answer could be scored as correct (Table~\ref{tab:S7}). Including forced answers retained at the 0.90 threshold, accuracy of retained answers fell to 91.3\% for GPT-6 Luna (21/23, from 95.5\%) and 78.6\% for Jev (11/14, from 84.6\%). When ``None of the above'' was added, GPT-6 Luna selected it in 4 of these 18 cases and none of the 52 with a correct listed option; Microsoft-Decision-1 selected it in 3 and 2, respectively, and Laya in 9 and 32. No other model selected it in any of the 18 cases. Across all 70 cases, accuracy was 61.4\% for GPT-6 Sol with medium reasoning, 51.4\% for GPT-6 Luna, 42.9\% for Clef, 35.7\% for Jev and 34.3\% for Microsoft-Decision-1. Among the locally run models, it ranged from 14.3\% (Laya) to 27.1\% (Gemma 4 12B). Adding None of the above also changed some answers in the 52 cases with a correct listed option. GPT-6 Sol with medium reasoning answered 43 of them correctly, against 40 with the listed options.

\begin{table}[htbp]
\caption{RheumBench: answers when the reference diagnosis is not among the options.}\label{tab:S7}
\centering\small
\setlength{\tabcolsep}{3pt}
\begin{tabular}{@{}>{\raggedright\arraybackslash\hangindent=1em\hangafter=1}p{4.2cm}cccccc@{}}
\toprule
 & \multicolumn{2}{c}{Listed option set} & \multicolumn{4}{c}{None of the above option set}\\
\cmidrule(lr){2-3}\cmidrule(l){4-7}
Model & \makecell[b]{Forced\\answers\\accepted at\\$\geq$0.90,\\n/18} & \makecell[b]{Accepted at\\$\geq$0.90 and\\correct,\\70 cases,\\n/N (\%)} & \makecell[b]{None of\\the above\\selected,\\n/18} & \makecell[b]{None of\\the above\\selected,\\n/52} & \makecell[b]{Correct\\with None\\of the above,\\n/52 (listed\\options)} & \makecell[b]{Accuracy with\\None of the\\above,\\70 cases,\\\% (95\% CI)}\\
\midrule
\multicolumn{7}{@{}l}{\textbf{\emph{System One models}}}\\
\quad Jev 1.13 & 1 & 11/14 (78.6) & 0 & 0 & 25 (26) & 35.7 (25.5--47.4)\\
\quad GPT-6 Luna (Decisions API) & 1 & 21/23 (91.3) & 4 & 0 & 32 (33) & 51.4 (40.0--62.8)\\
\quad Clef & 0 & 5/5 (100.0) & 0 & 0 & 30 (31) & 42.9 (31.9--54.5)\\
\quad Microsoft-Decision-1 & 1 & 11/16 (68.8) & 3 & 2 & 21 (21) & 34.3 (24.2--46.0)\\
\quad Laya 0.3.20 & 7 & 1/20 (5.0) & 9 & 32 & 1 (5) & 14.3 (7.9--24.3)\\
\addlinespace[3pt]
\multicolumn{7}{@{}l}{\textbf{\emph{Primary comparator}}}\\
\quad GPT-6 Sol (medium) & 1 & 23/24 (95.8) & 0 & 0 & 43 (40) & 61.4 (49.7--72.0)\\
\addlinespace[3pt]
\multicolumn{7}{@{}l}{\textbf{\emph{Secondary comparators}}}\\
\quad GPT-6 Sol (none) & 0 & 15/15 (100.0) & 0 & 1 & 41 (39) & 58.6 (46.9--69.4)\\
\quad Gemma 4 12B (4-bit, no thinking) & 0 & 1/1 (100.0) & 0 & 0 & 19 (21) & 27.1 (18.1--38.5)\\
\quad Qwen3-8B (4-bit, no thinking) & 1 & 0/1 (0.0) & 0 & 1 & 18 (16) & 25.7 (16.9--37.0)\\
\bottomrule
\end{tabular}
\tabnote{%
Exploratory; run 1. Listed option set (second and third columns): in the 18 cases whose reference diagnosis is not among the options every answer is wrong under the answer key, and the third column counts these forced answers as errors among all answers accepted at $\geq$0.90 over the 70 cases. System One probabilities are normalised over the options supplied, so a forced answer can receive a high probability as the least implausible option. None of the above option set (last four columns): None of the above was added as the last option of every case and is the correct answer of the 18 cases; zero selections in 18 cases do not show that a model never recognises a missing diagnosis. Adding None of the above changed every request, so answers to the 52 cases with a correct listed option could change; the number correct with the listed options (Table~\ref{tab:1} of the main text) is given in parentheses. Accuracy with Wilson 95\% CIs. Laya's 70-case accuracy reflects indiscriminate selection of None of the above.
}
\end{table}

\section{RheumaMIR: errors with the highest probabilities}\label{sec:S7}

Table~\ref{tab:S8} shows the ten RheumaMIR questions answered incorrectly with the highest selected-answer probabilities by the main models in run 1. In all ten, the highest-probability error was Microsoft-Decision-1's in Spanish (0.967--0.999); Microsoft-Decision-1 selected the same wrong option in English in six of them, with a probability of at least 0.90 in two. Five questions asked for the next investigation, three for treatment or prevention, and two, both negatively worded, for the incorrect statement or the atypical sign. Jev or Clef selected the same wrong option in six of the questions, but with a probability of at least 0.90 only in question 33, which Jev, Clef and Microsoft-Decision-1 answered incorrectly in both languages. GPT-6 Sol with medium reasoning and GPT-6 Luna answered all ten correctly.

{\footnotesize
\setlength{\tabcolsep}{4pt}
\setlength{\LTpost}{0pt}
\begin{longtable}{@{}>{\raggedright\arraybackslash}p{2.25cm}>{\raggedright\arraybackslash}p{6.65cm}>{\raggedright\arraybackslash}p{3.1cm}>{\raggedright\arraybackslash}p{3.6cm}@{}}
\caption{RheumaMIR: the ten questions answered incorrectly with the highest selected-answer probabilities.}\label{tab:S8}\\
\toprule
ID & Question (abridged) and options & Incorrect option selected & Model, language: probability\\
\midrule
\endfirsthead
\multicolumn{4}{@{}l}{\small\textit{Table~\ref{tab:S8} (continued)}}\\
\toprule
ID & Question (abridged) and options & Incorrect option selected & Model, language: probability\\
\midrule
\endhead
\bottomrule
\endfoot
\textbf{138}\par 2010--2011\par Clinical case\par SLE & \emph{Woman with SLE on glucocorticoids (0.5 mg/kg/day for several months); sudden, continuous knee pain for 3 weeks; normal radiograph. Next investigation?}\par 1. Knee CT scan.\par 2. Full blood count with a differential white cell count, and C-reactive protein.\par 3. Cell count and biochemistry of joint fluid.\par 4. Culture of joint fluid.\par \keyopt{\textbf{5. Magnetic resonance imaging.}} & 3. Cell count and biochemistry of joint fluid. & Microsoft-Decision-1, Spanish: \textbf{0.999}\par Microsoft-Decision-1, English: \textbf{0.966}\par Clef, Spanish: 0.827\\
\addlinespace[4pt]
\textbf{151}\par 2023--2024\par Clinical case\par Bone metabolism & \emph{93-year-old man after hemiarthroplasty for an osteoporotic subcapital femoral fracture. Correct statement?}\par 1. He is not a candidate for bone-protective treatment because of his age and dementia.\par \keyopt{\textbf{2. Bisphosphonates should not be started if he has hypocalcaemia.}}\par 3. He should receive calcium supplements even if his levels are normal.\par 4. He should remain on bed rest for six weeks after surgery to improve fracture healing. & 3. He should receive calcium supplements even if his levels are normal. & Microsoft-Decision-1, Spanish: \textbf{0.997}\par Microsoft-Decision-1, English: 0.772\par Clef, Spanish: 0.463\\
\addlinespace[4pt]
\textbf{63}\par 2015--2016\par Clinical case\par SLE & \emph{37-year-old man with MCP arthritis, pleuritis and malar erythema; ANA 1/640, anti-DNA positive, anti-Sm negative. Initial treatment of choice?}\par 1. High-dose glucocorticoids.\par 2. Glucocorticoids and mycophenolate.\par \keyopt{\textbf{3. Non-steroidal anti-inflammatory drugs and antimalarials.}}\par 4. The condition is likely to be self-limiting and does not require treatment. & 1. High-dose glucocorticoids. & Microsoft-Decision-1, Spanish: \textbf{0.993}\par Microsoft-Decision-1, English: 0.829\\
\addlinespace[4pt]
\textbf{142}\par 2009--2010\par Clinical case\par Others & \emph{72-year-old man with macroglossia, periorbital ecchymosis, sensorimotor polyneuropathy, monoclonal peak, proteinuria 6 g/24 h and prothrombin activity 35\%. Diagnostic approach?}\par \keyopt{\textbf{1. Abdominal fat biopsy and bone marrow aspirate.}}\par 2. Determination of microalbuminuria and glycated haemoglobin.\par 3. Determination of plasma vitamin B12 and erythrocyte folate.\par 4. Serology for HIV (human immunodeficiency virus).\par 5. Renal biopsy. & 5. Renal biopsy. & Microsoft-Decision-1, Spanish: \textbf{0.990}\par Microsoft-Decision-1, English: 0.860\\
\addlinespace[4pt]
\textbf{161}\par 2024--2025\par Clinical case\par SLE & \emph{Woman with SLE diagnosed 1 month earlier (autoimmune haemolytic anaemia, prolonged aPTT), on glucocorticoids, hydroxychloroquine and belimumab; dyspnoea for 48 h, haemoglobin 10.2 g/dL. First laboratory test?}\par \keyopt{\textbf{1. D-dimer.}}\par 2. Urine protein.\par 3. NT-proBNP.\par 4. Coombs' test. & 4. Coombs' test. & Microsoft-Decision-1, Spanish: \textbf{0.986}\par Jev, English: 0.620\par Jev, Spanish: 0.550\\
\addlinespace[4pt]
\textbf{33}\par 2019--2020\par Factual\par Vasculitis & \emph{With regard to temporal arteritis, mark the WRONG answer.}\par 1. It occurs mainly in white people over the age of 50 years.\par 2. Approximately a quarter of patients develop aneurysm or dilatation of the thoracic aorta.\par 3. Unilateral or bilateral vision loss is one of the most serious complications.\par \keyopt{\textbf{4. The peak incidence is between the sixth and seventh decades of life.}} & 2. Approximately a quarter of patients develop aneurysm or dilatation of the thoracic aorta. & Microsoft-Decision-1, Spanish: \textbf{0.984}\par Clef, English: \textbf{0.980}\par Clef, Spanish: \textbf{0.960}\par Microsoft-Decision-1, English: \textbf{0.948}\par Jev, English: \textbf{0.940}\par Jev, Spanish: \textbf{0.900}\\
\addlinespace[4pt]
\textbf{42}\par 2018--2019\par Factual\par Spondyloarthritis & \emph{Which of the following radiological signs is NOT typical of ankylosing spondylitis?}\par 1. Squaring of the vertebral bodies.\par \keyopt{\textbf{2. Calcification of the anterior longitudinal ligament.}}\par 3. Sacroiliitis.\par 4. Thoracic kyphosis. & 4. Thoracic kyphosis. & Microsoft-Decision-1, Spanish: \textbf{0.981}\par Clef, English: 0.847\par Clef, Spanish: 0.757\par Jev, Spanish: 0.670\par Jev, English: 0.600\\
\addlinespace[4pt]
\textbf{78}\par 2013--2014\par Clinical case\par Bone metabolism & \emph{75-year-old woman with temporal arteritis about to start high-dose prednisone for at least 1 year; DXA waiting time 4--5 months. Osteoporosis prevention?}\par 1. Request DXA and wait for the result.\par 2. Assess absolute risk of fracture by means of the FRAX questionnaire without BMD and treat only if it is high.\par \keyopt{\textbf{3. Start treatment with bisphosphonates and vitamin D (800 IU/day).}}\par 4. Administer calcium supplement (1 g) and vitamin D (800 IU/day).\par 5. Assess for osteopenia on X-rays and treat if present. & 4. Administer calcium supplement (1 g) and vitamin D (800 IU/day). & Microsoft-Decision-1, Spanish: \textbf{0.973}\par Microsoft-Decision-1, English: 0.517\\
\addlinespace[4pt]
\textbf{132}\par 2019--2020\par Clinical case\par Infectious arthritis & \emph{70-year-old man with progressive low back pain for 1 week, marked limitation of spinal mobility and low-grade fever; normal lumbar radiograph. Diagnosis to RULE OUT first, and technique?}\par 1. Vertebral compression fracture with computed tomography (CT).\par 2. Bone metastases with technetium bone scintigraphy.\par 3. Ankylosing vertebral hyperostosis with thoracolumbar spine radiography.\par \keyopt{\textbf{4. Pyogenic spondylodiscitis with magnetic resonance imaging (MRI).}} & 3. Ankylosing vertebral hyperostosis with thoracolumbar spine radiography. & Microsoft-Decision-1, Spanish: \textbf{0.969}\\
\addlinespace[4pt]
\textbf{32}\par 2020--2021\par Clinical case\par Vasculitis & \emph{37-year-old man with asthenia, low-grade fever, weight loss, polyneuropathy and palpable purpura; normal blood pressure, renal function and urine sediment. Investigation to confirm the most likely diagnosis?}\par 1. Doppler ultrasound of the renal arteries for the diagnosis of Takayasu's arteritis.\par 2. A punch skin biopsy for the diagnosis of polyarteritis nodosa.\par \keyopt{\textbf{3. Detection of circulating cryoglobulins for the diagnosis of cryoglobulinaemic vasculitis.}}\par 4. A renal biopsy for diagnosis of IgA vasculitis. & 2. A punch skin biopsy for the diagnosis of polyarteritis nodosa. & Microsoft-Decision-1, Spanish: \textbf{0.967}\par Jev, English: 0.800\par Jev, Spanish: 0.710\par Clef, Spanish: 0.548\\
\addlinespace[4pt]
\end{longtable}
}
\vspace{-\baselineskip}
\tabnote{%
Run 1. Questions are ranked by the highest selected-answer probability given to an incorrect answer by any of the main models (Jev 1.13, GPT-6 Luna [Decisions API], Clef, Microsoft-Decision-1 and GPT-6 Sol with medium reasoning) in either language; the eleventh, question 141 (0.966), was within 0.001 of the tenth. ID, dataset ID, with examination call, question type and disease category. Questions are abridged; options are given in English with their original numbers, which are the same in Spanish, and the full text of each question is in the deposited dataset. The keyed (correct) option is in bold on a green background. Every incorrect answer of the main models to these questions is listed with its selected-answer probability; probabilities of at least 0.90, which the threshold would have retained, are in bold. In each question, every model that answered incorrectly selected the same option. GPT-6 Sol with medium reasoning and GPT-6 Luna answered all ten questions correctly in both languages. Jev returns probabilities rounded to two decimals. Laya and the local generative models are not shown. SLE, systemic lupus erythematosus.
}

\section{Reproducibility}\label{sec:S8}

\textbf{Code and data.} For each benchmark, an evaluation notebook sends the requests and saves every request and response, and an analysis notebook computes the reported tables and figures from the saved responses without calling any model or requiring credentials. All scores, extracted answers and probability vectors were checked against the raw responses, and the content of every request against the dataset files. The dataset, the code and the results for RheumaMIR are attached as \textbf{SM\_code\_dataset\_statistics.} The RheumBench vignettes belong to the RheumBench study. The dataset, the code and the results for RheumBench will be made available by JK upon reasonable request.

\end{document}